\RequirePackage{fix-cm}
\documentclass[smallextended]{svjour3}       
\smartqed  

\usepackage{subfigure}
\usepackage{epstopdf}
\usepackage{epsfig}
\usepackage{amsmath}
\usepackage{graphicx}
\usepackage{graphicx}
\usepackage{threeparttable}
\usepackage{float}
\usepackage{caption}
\usepackage{booktabs}
\usepackage{marvosym}

\usepackage{colortbl}
\usepackage{color}
\usepackage{soul}

\usepackage{array}
\usepackage{amsmath,amssymb,amsfonts}
\usepackage{threeparttable}
\usepackage[T1]{fontenc}
\usepackage{lineno,hyperref}

\usepackage{multirow}
\newtheorem{myDef}{Definition}
\usepackage{indentfirst}
\begin{document}

\title{Wide Aspect Ratio Matching for Robust Face Detection}


\author{Shi Luo$^{1,2}$ \and Xiongfei Li$^{1,2}$ \and Xiaoli Zhang$^{1,2}$
}

\institute{{\small  \Letter{\small }}~Xiaoli Zhang \at
                 ~\email{zhangxiaoli@jlu.edu.cn}
                 \and
           $^{1}$ ~Key Laboratory of Symbolic Computation and Knowledge Engineering of Ministry of Education, Jilin University, Changchun 130012, China\\
           \\
           $^{2}$ ~College of Computer Science and Technology, Jilin University, Changchun 130012, China
        }

\date{Received: date / Accepted: date}

\maketitle

\begin{abstract}
Recently, anchor-based methods have achieved great progress in face detection. Once anchor design and anchor matching strategy determined, plenty of positive anchors will be sampled. However, faces with extreme aspect ratio always fail to be sampled according to standard anchor matching strategy. In fact, the max IoUs between anchors and extreme aspect ratio faces are still lower than fixed sampling threshold. In this paper, we firstly explore the factors that affect the max IoU of each face in theory. Then, anchor matching simulation is performed to evaluate the sampling range of face aspect ratio. Besides, we propose a Wide Aspect Ratio Matching (WARM) strategy to collect more representative positive anchors from ground-truth faces across a wide range of aspect ratio. Finally, we present a novel feature enhancement module, named Receptive Field Diversity (RFD) module, to provide diverse receptive field corresponding to different aspect ratios. Extensive experiments show that our method can help detectors better capture extreme aspect ratio faces and achieve promising detection performance on challenging face detection benchmarks, including WIDER FACE and FDDB datasets.
\end{abstract}

\section{Introduction}
\label{sec1}
Accurate face detection is a prerequisite of many face related applications, such as face alignment \cite{PIFA,LAB}, face recognition \cite{SphereFace,CosFace,ArcFace} and facial expression recognition, etc. In recent years, convolutional neural networks (CNN) have achieved remarkable successes in face detection. Unlike traditional face detectors of hand-crafted features, CNN-based methods utilize popular backbone network(e.g. VGG \cite{VGG16}, ResNet \cite{ResNet}, DenseNet \cite{Densenet}, MobileNet) and feature fusion strategy(e.g. FPN \cite{FPN}) to extract robust face features automatically.\\

\indent{}Among them, anchor-based methods play a dominant role in CNN-based face detectors. During the training phase, a series of anchors are preset on the images with different scales and single aspect ratio (AR). Later, standard anchor matching (SAM) strategy \cite{FasterR-CNN} assigns positive and negative labels on these anchors according to intersection-over-union (IoU) threshold. We assign anchors to positive labels when their IoU threshold are in ($T_{p}$, 1], and negative labels if their IoU in [0, $T_{n}$). The rest anchors are discarded. Finally, these labeled anchors are feed into training network and update its parameters. During the inference phase, they detect faces by classifying and regressing these anchors. Predict score from classification branch can tell us if current anchor contain a face while coordinate offsets of current anchor locates its position from regression branch. \\

\begin{figure}[htbp]
		\centering
		\subfigure[0.426966 ]{
			\begin{minipage}[t]{0.185\linewidth}
				\centering
				\includegraphics[width=1.025\columnwidth]{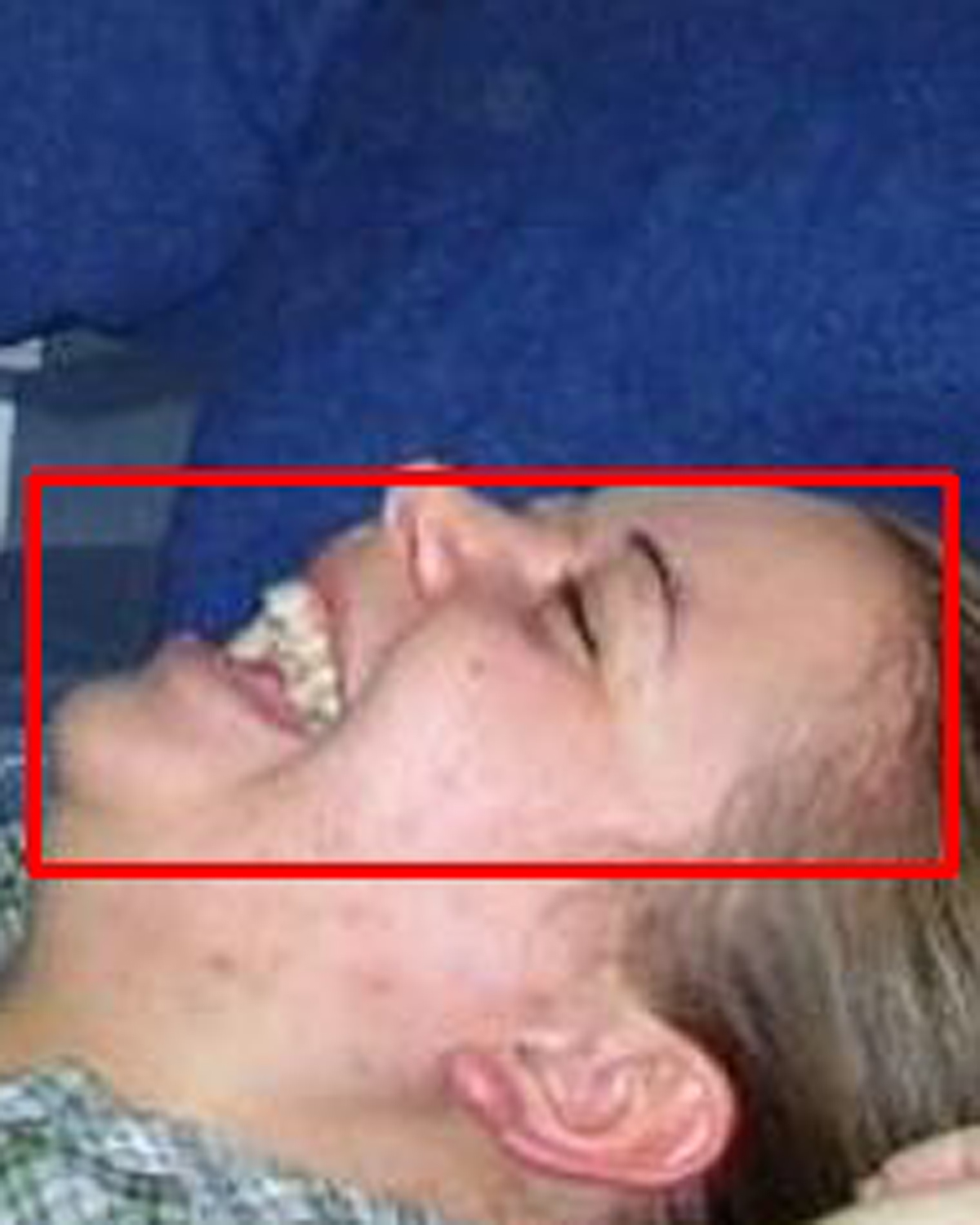}
			\end{minipage}%
		}%
		\subfigure[2.368421 ]{
			\begin{minipage}[t]{0.185\linewidth}
				\centering
				\includegraphics[width=1.025\columnwidth]{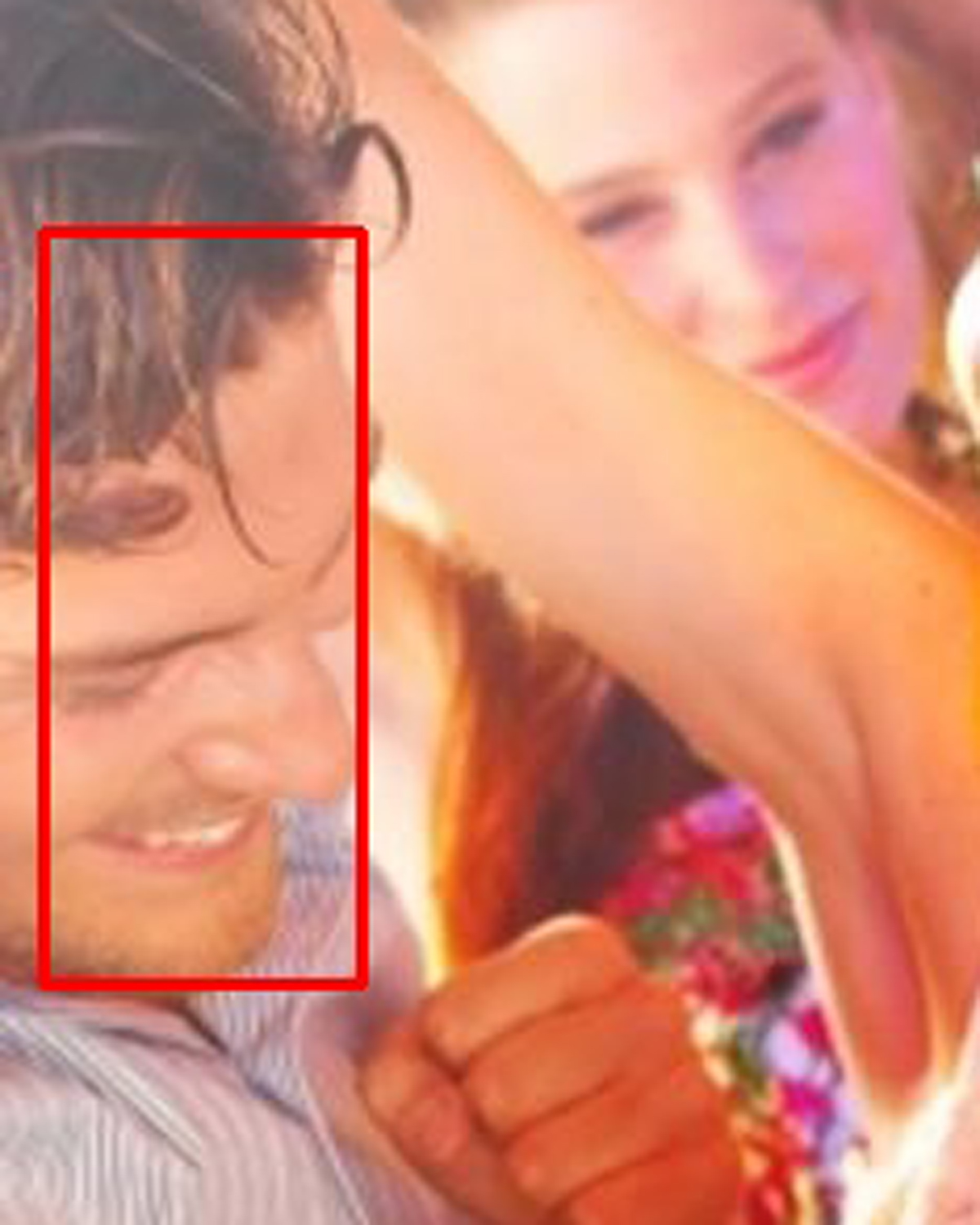}
			\end{minipage}%
		}%
		\subfigure[2.396825]{
			\begin{minipage}[t]{0.179\linewidth}
				\centering
				\includegraphics[width=1.065\columnwidth]{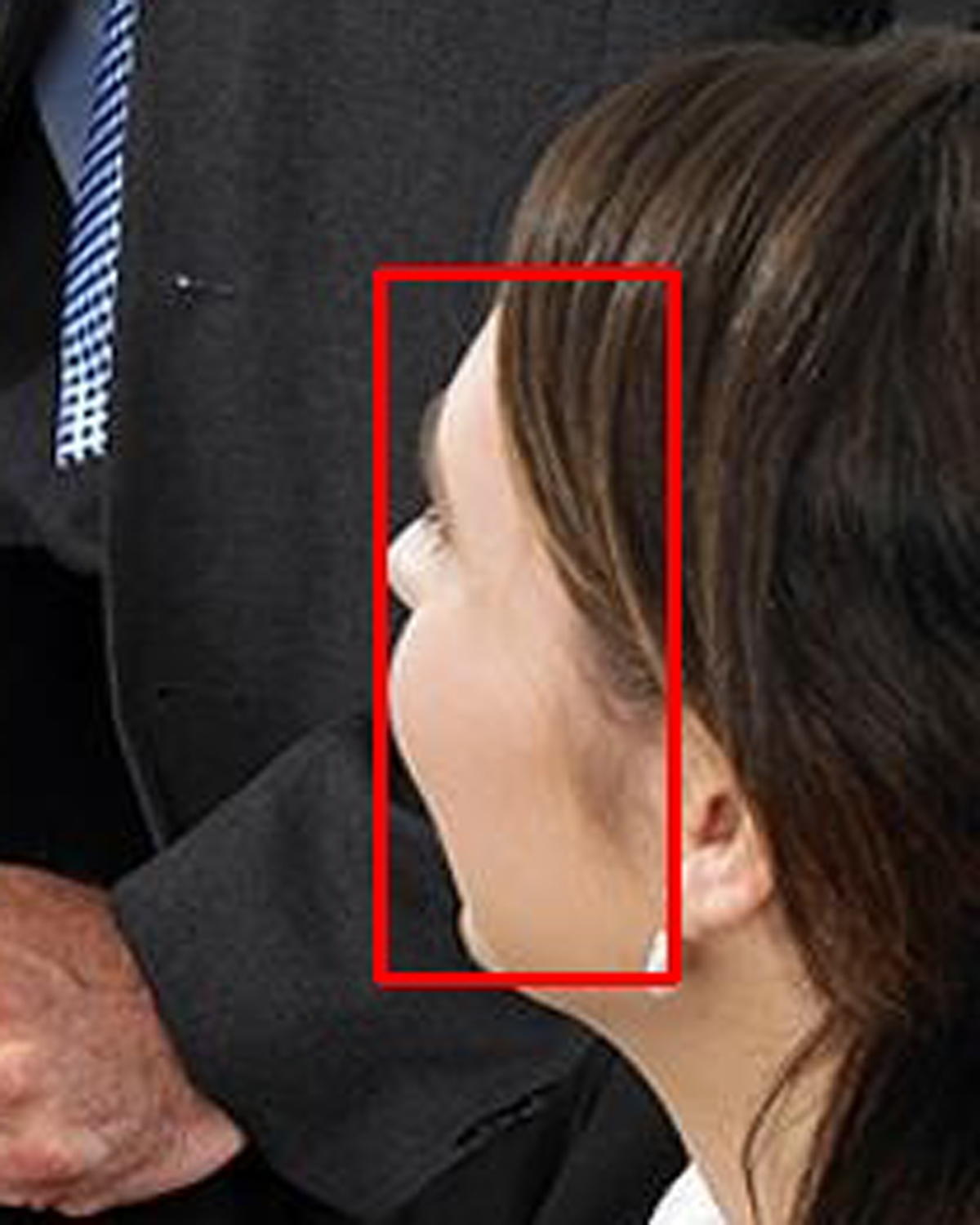}
			\end{minipage}
		}%
		\subfigure[2.482759]{
			\begin{minipage}[t]{0.179\linewidth}
				\centering
				\includegraphics[width=1.065\columnwidth]{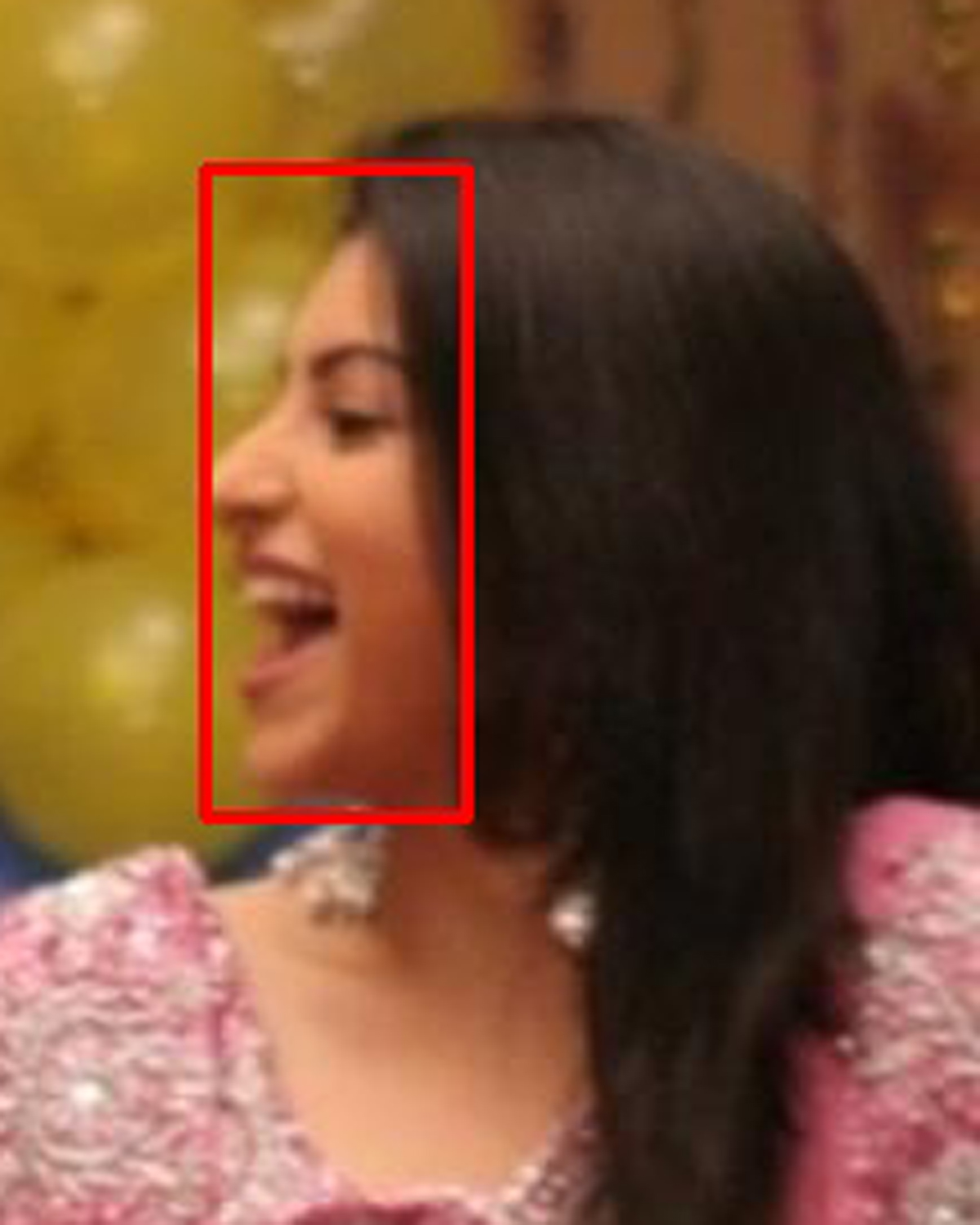}
			\end{minipage}
		}%
		\subfigure[2.400000]{
			\begin{minipage}[t]{0.185\linewidth}
				\centering
				\includegraphics[width=1.025\columnwidth]{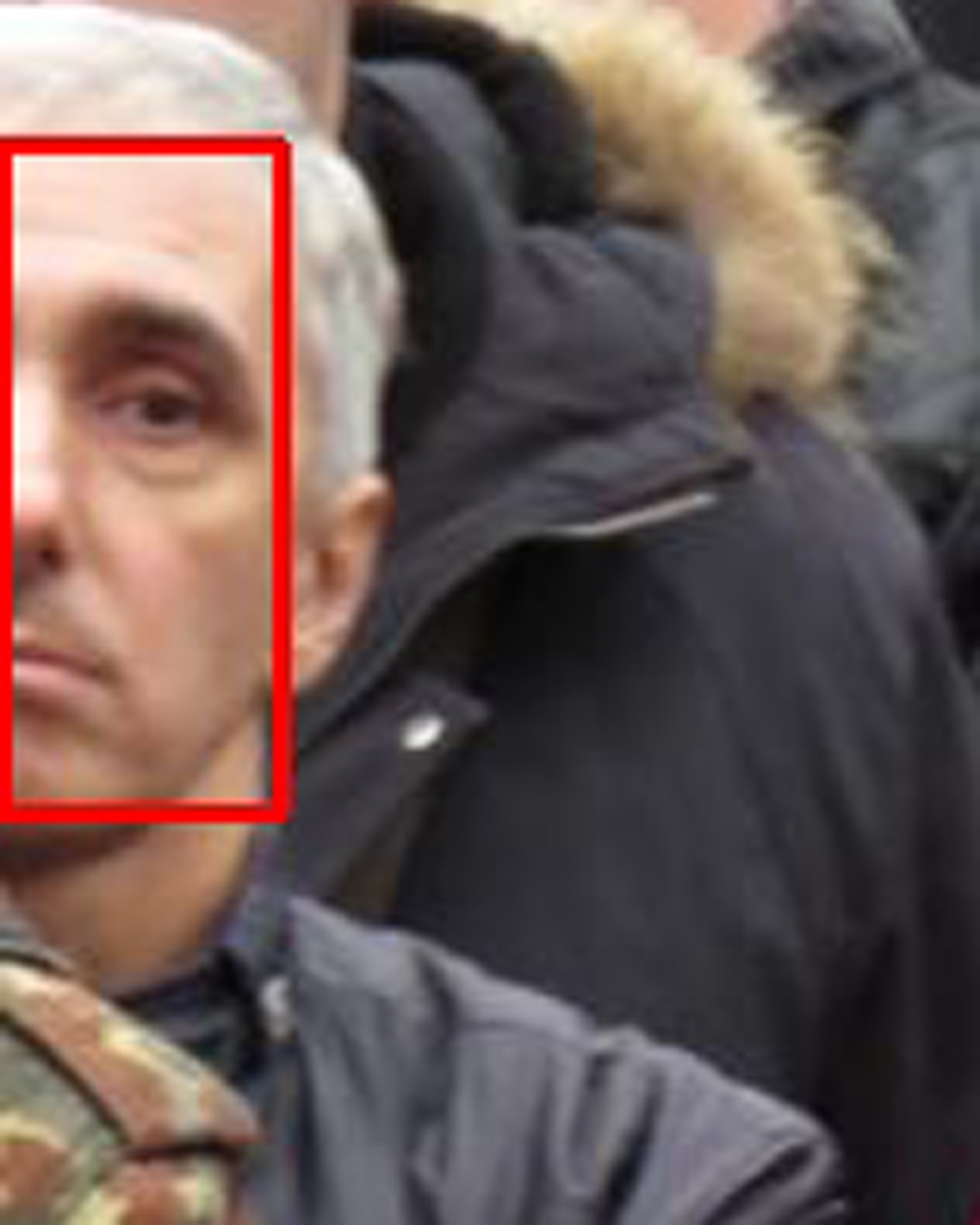}
			\end{minipage}%
		}%
		\\
		\subfigure[0.449275]{
			\begin{minipage}[t]{0.185\linewidth}
				\centering
				\includegraphics[width=1.025\columnwidth]{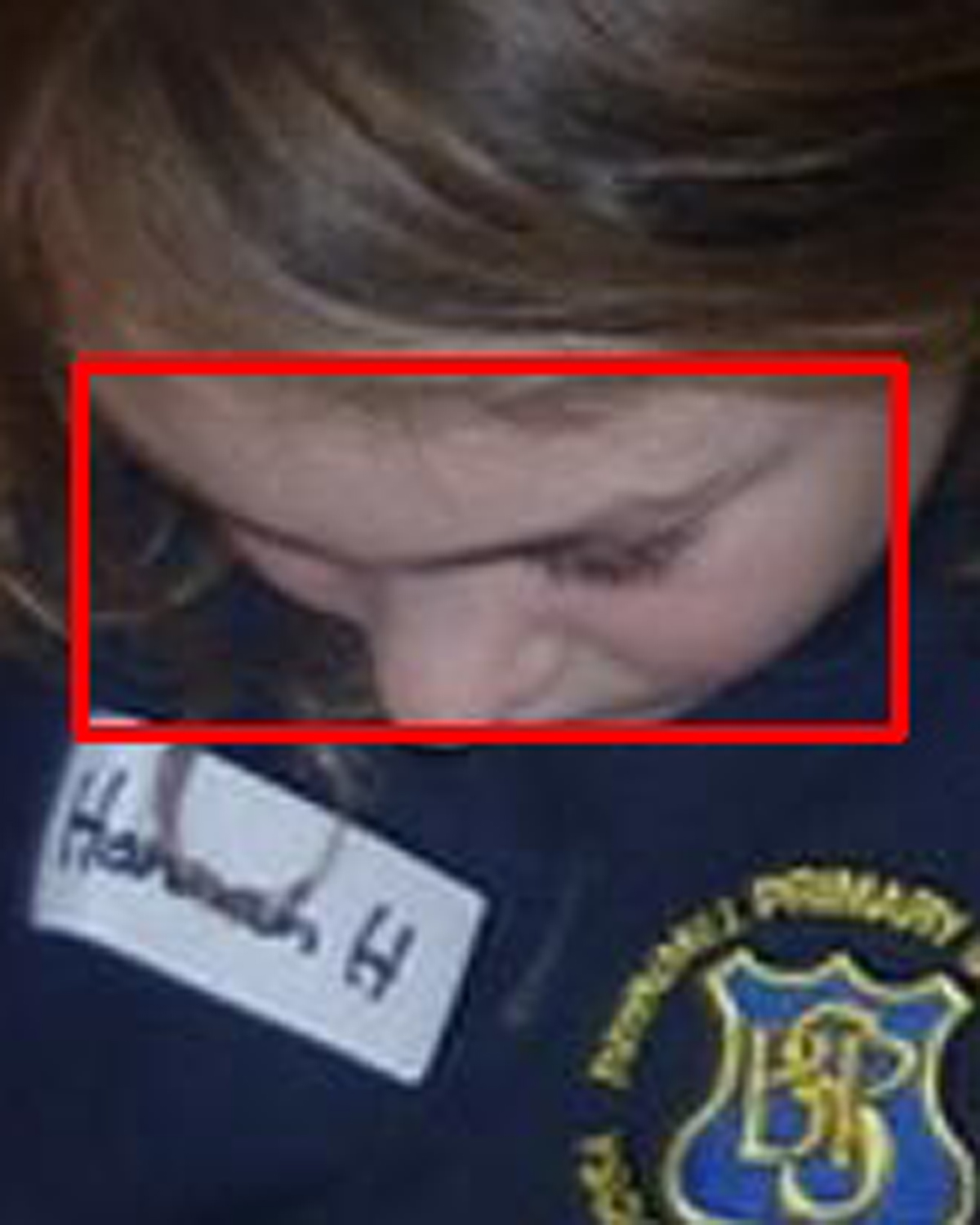}
			\end{minipage}%
		}%
		\subfigure[0.488889]{
			\begin{minipage}[t]{0.185\linewidth}
				\centering
				\includegraphics[width=1.025\columnwidth]{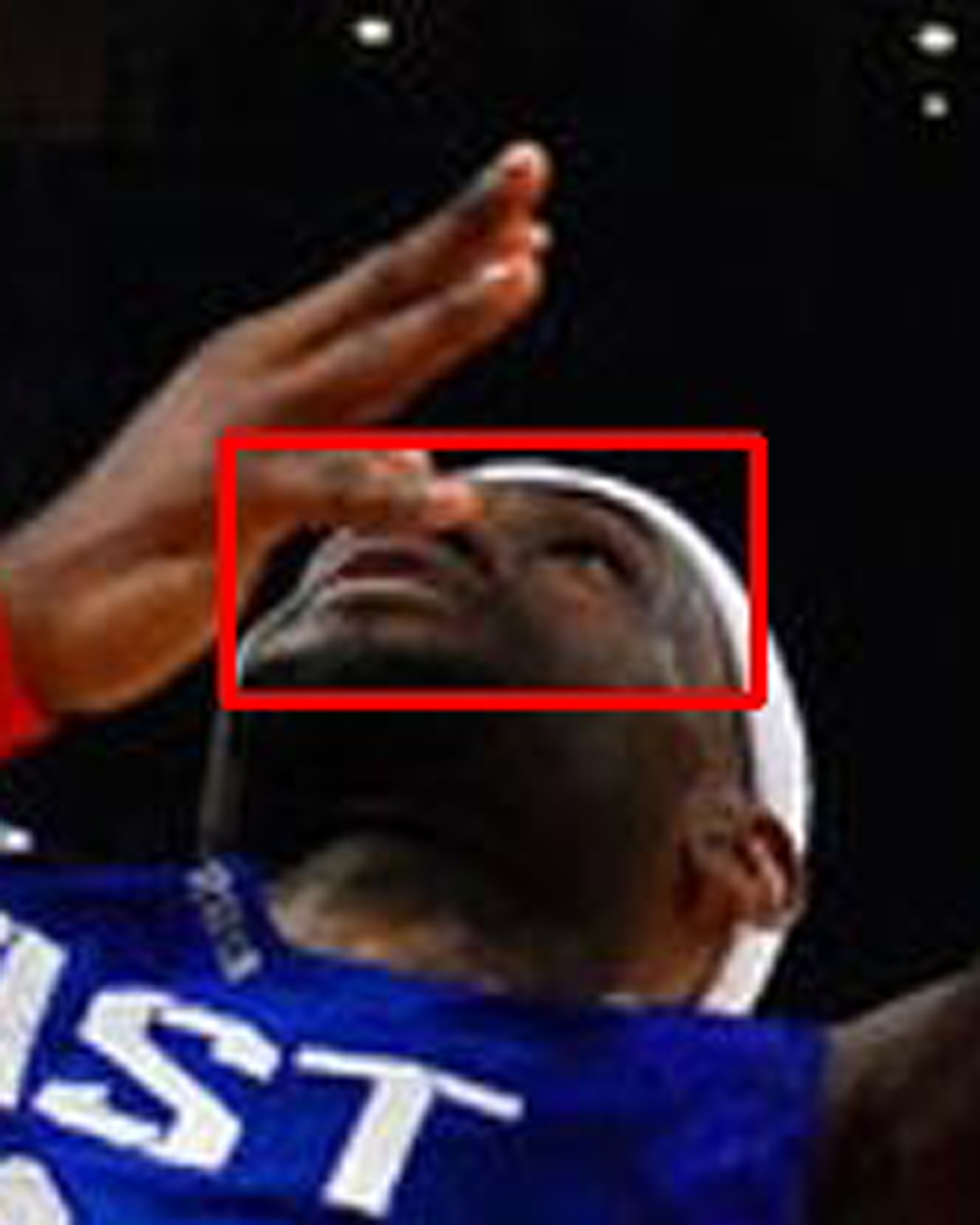}
			\end{minipage}%
		}%
		\subfigure[0.538922]{
			\begin{minipage}[t]{0.178\linewidth}
				\centering
				\includegraphics[width=1.065\columnwidth]{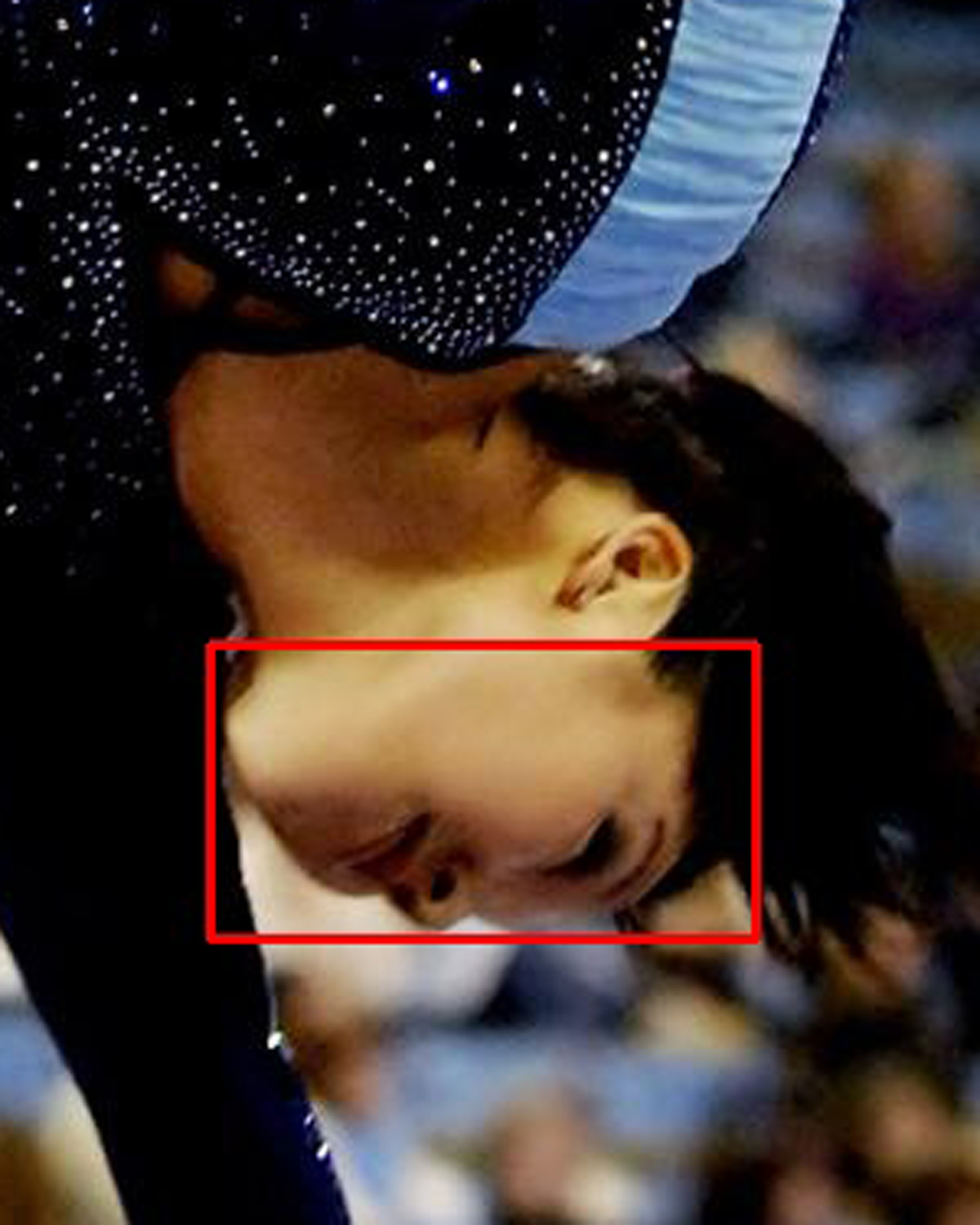}
			\end{minipage}
		}%
		\subfigure[3.051948]{
			\begin{minipage}[t]{0.185\linewidth}
				\centering
				\includegraphics[width=1.025\columnwidth]{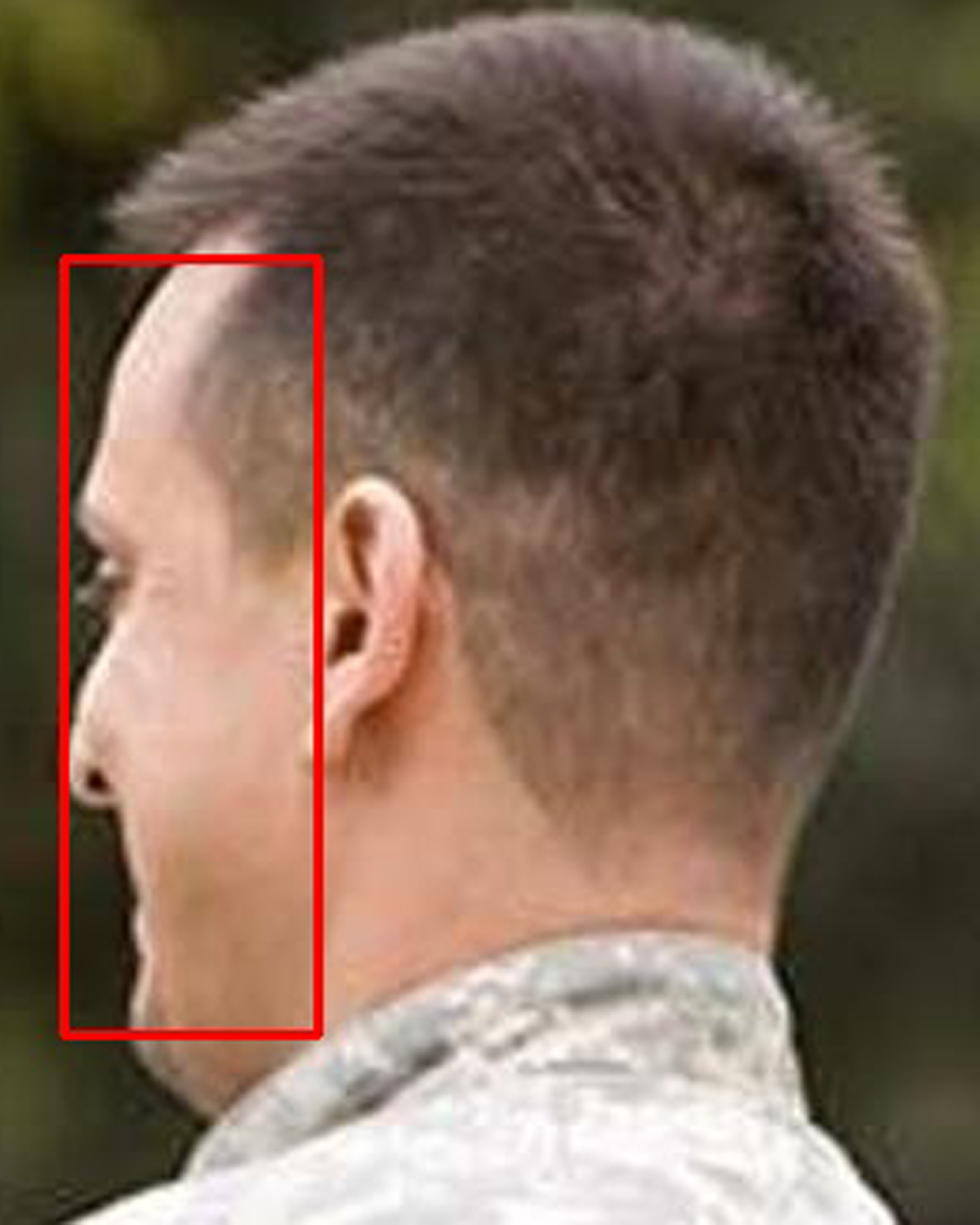}
			\end{minipage}%
		}%
		\subfigure[0.470270]{
			\begin{minipage}[t]{0.185\linewidth}
				\centering
				\includegraphics[width=1.025\columnwidth]{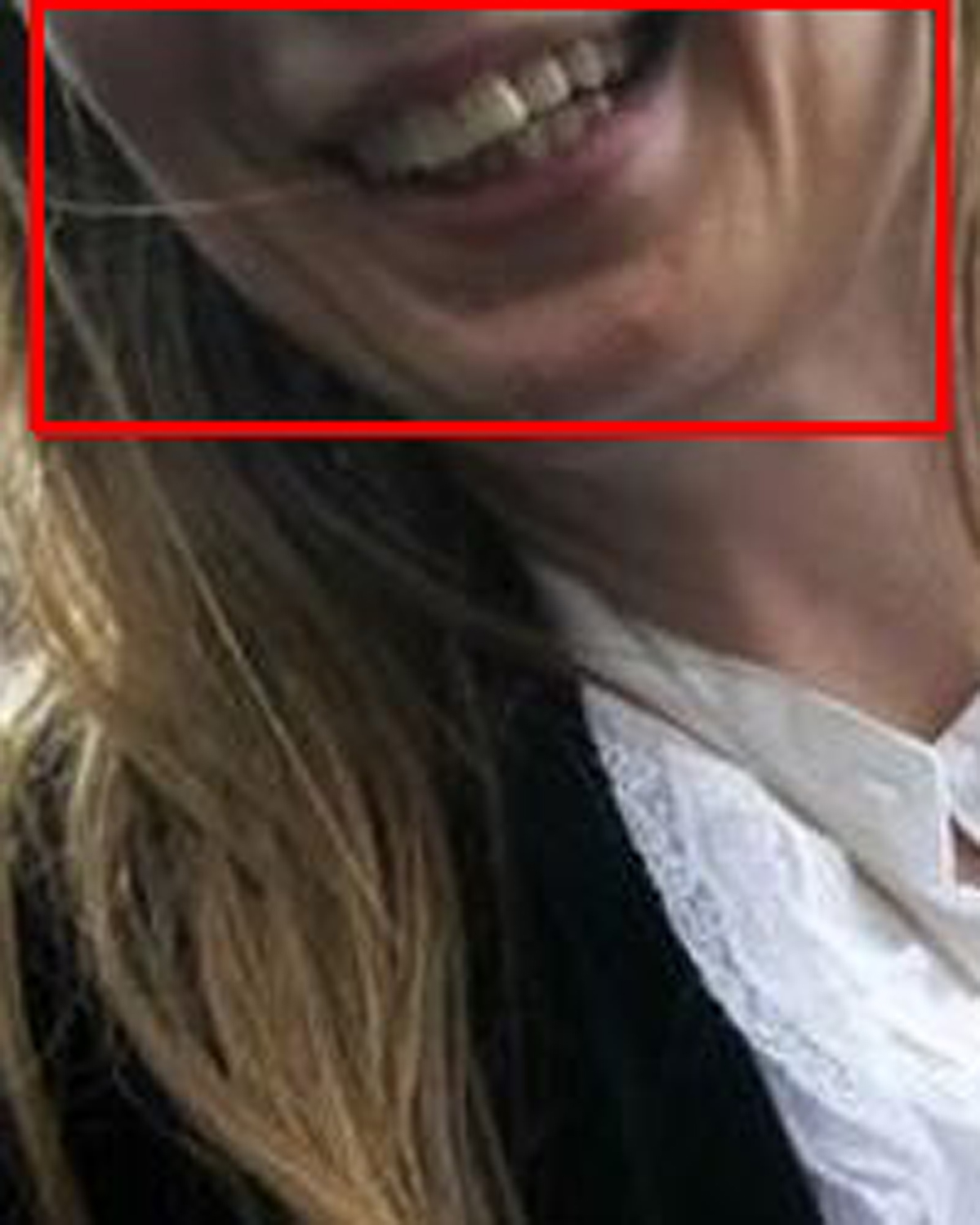}
			\end{minipage}%
		}%
		\centering
		\caption{Extreme aspect ratio faces on Wider Face training set. The aspect ratio (AR) of each ground truth face is noted blow. The first row faces fail to be sampled with anchor AR of 1.0 while second row faces with anchor AR of 1.25. The first four columns show extreme pose faces while the last column are partial faces.
		\label{fig1}}
	\end{figure}

\indent{}However, sampling positve anchors from each face is not always successful. SAM strategy utilizes the identical sampling threshold for all faces. When anchor sampling threshold fixed, the AR sampling range for positive anchors is determined. The ground-truth faces whose AR out of this range, called extreme aspect ratio faces, will be neglect. Figure \ref{fig1} shows some extreme AR faces on Wider Face \cite{WIDERFACE} training set. We can see that extreme AR faces consist of extreme pose faces and partial faces. Extreme pose faces come from head rotation in pitch, yaw, and roll axis (see (a)-(d) and (f)-(i)). And partial faces are derived from occlusion (see (e), (j)). For sampling more positive anchors related to extreme AR faces, fixed sampling threshold for all faces is no longer suitable.   

\indent{}Furthermore, diversified aspect raito of faces conflict with single receptive field of network. Current feature enhancement modules usually adopt square filter to enlarge receptive field. In fact, anchors with diverse receptive field make it easier to be classified and regressed in a convolutional manner.

\indent{}In this paper, we firstly investigate what makes the max IoU of each face different and theoretically prove that AR of faces is the key impact. Then, anchor matching simulation is performed to evaluate the sampling range of face AR. Obviously, the failure of sampling extreme AR faces is just because of their AR out of sampling range. In fact, the max IoUs of these extreme AR faces are still lower than fixed sampling threshold in SAM strategy. Therefore, we propose a Wide Aspect Ratio Matching (WARM) strategy to collect more representative positive anchors from ground-truth faces with a wide range of AR. Specifically, extreme AR faces have their own sampling threshold according to AR themselves. Besides, we design a novel feature enhancement module, named Receptive Field Diversity (RFD) module, to provide more diverse aspect ratio and large receptive field simultaneously. Both symmetric and asymmetric convolution kernels are used. 
Finally, we conduct comprehensive experiments on popular benchmarks, including WIDER FACE and FDDB \cite{FDDB} datasets, and achieve promising detection performance, especially for extreme AR faces.\\
\indent{}For clarity, the main contributions of this paper can be summarized as:\\
\indent{}(1) We theoretically prove that aspect ratio of faces is the key factor affecting the max IoU of each face and perform anchor matching simulation to evaluate the sampling range of face aspect ratio.\\
\indent{}(2) We propose a WARM strategy to collect more representative positive anchors from ground-truth faces with a wide range of aspect ratio.\\
\indent{}(3) A novel feature enhancement module, named RFD module, is designed to provide more diverse aspect ratio and large receptive field simultaneously.\\
\indent{}(4) Our method can help detectors better capture extreme aspect ratio faces and achieve promising detection performance on challenging face detection benchmarks, including WIDER FACE and FDDB datasets.\\

\indent{}The rest of the paper is organized as follows. Section \ref{sec2} briefly reviews the related work in face detection. Section \ref{sec3} presents our proposed method. Section \ref{sec4} shows our experimental results. Section \ref{sec5} concludes this paper.

\section{Related work}
\label{sec2}
Face detection is a fundamental step to various face related applications. Previous methods can be roughly divided into two categories as follows:\\
\indent Traditional method: They detect faces based on hand-crafted features in a sliding-window manner and optimize each component separately. The pioneering work of Viola-Jones \cite{VJ} uses Haar-like features and AdaBoost to train a cascade of face detectors. LBP \cite{LBP} introduces local texture features for face detection. NPDFace \cite{NPDFace} proposes normalized pixel difference feature to address challenges in unconstrained face detection, such as arbitrary pose and heavy occlusion. Sawat et al. \cite{Pixelencoding} adpots hand-crafted as well as visual features together for improving robustness in face detection. \\

\indent CNN-based method: Different from traditional methods, CNN-based methods can automatically extract discriminative features from challenging face datasets. CascadeCNN \cite{CascadeCNN} developes a cascade framework via deep CNNs to detect face coarse to fine. Faceness \cite{Faceness} trains a series of CNNs for facial attribute recognition to detect partially occluded faces. MTCNN \cite{MTCNN} jointly solves face detection and alignment using several multi-task CNNs. HR \cite{HR} builds multi-level image pyramids to boost the performance on extreme scale variations. UnitBox \cite{UnitBox} presents an IoU loss to directly regress the bounding box. FANet \cite{FANet} creates a new hierarchical effective feature pyramid with rich semantics at all scales. BFBox \cite{BFBox} designs a FPN-attention module to joint search the face-appropriate space of backbone and FPN. \\

Recently, anchor-based methods have attracted more attention duo to their detection accuracy as well as inference efficiency. Effective anchor design and anchor matching strategy are necessary to generate more representative training samples. Besides, feature enhancement can further improve the ability of discriminative face features. Therefore, we focus on reviewing the prior works from this three perspectives below: \\

\indent \textbf{Anchor Design:} In face detection, the choice of anchors and their placement on the image is very important. For example, using extra strided anchors are shown to be beneficial. ZCC \cite{Zhu} introduces a novel anchor design to guarantee high overlaps between faces and anchor boxes. PyramidBox \cite{Pyramidbox} formulates a data-anchor-sampling strategy to increase the proportion of small faces in the training data. FaceBoxes \cite{FaceBoxes} presents a new anchor densification strategy to improve the recall rate of small faces. FA-RPN \cite{FA-RPN} proposes an efficient anchor placement strategy to reduce the number of anchors to detect faces. Group sampling \cite{GroupSampling} emphasizes the importance of balanced training samples, including both positive and negative ones, at different scales. In this paper, we continue to follow these guidances to form a high recall ratio anchor design, which adopts a wider range of anchor size and a shorter anchor stride. \\

\indent \textbf{Anchor Matching:} SAM strategy  utilizes fixed sampling threshold to assign positive anchors. S$^{3}$FD \cite{S3FD} proposes scale compensation anchor matching strategy which helps the outer faces match more anchors. SRN \cite{SRN} introduces a selective two-step classification to ignore training easy sample anchors in the second stage. DSFD \cite{DSFD} offers an improved anchor matching method to provide better initialization for the regressor. HAMBox \cite{HAMBox} helps outer faces compensate high-quality anchors, which can obtain high IoU regression bounding boxes. Although anchor matching has being extensively studied, the failure of sampling positive anchors from extreme AR faces is still neglect. In this paper, we propose WARM strategy to sample more positive anchors from extreme AR faces. \\

\indent \textbf{Feature Enhancement:} SSH \cite{SSH} adds large filters on each detection module to merge the context information. DSFD \cite{DSFD} introduces a feature enhance module to extend the single shot detector to dual shot detector. OS-LFFD \cite{OS-LFFD} design a novel ommateum block to maintain the corresponding ratio of receptive fields to face regions. RefineFace \cite{RefineFace} construct a RFE module to provide more diverse receptive fields for detecting extreme-pose faces. In this paper, we propose RFD module to adapt the diversity of face AR. Inspired by ACNet \cite{ACNet}, both symmetric and asymmetric convolution kernels are used into proposed RFD module. \\

\section{Proposed Method}
\label{sec3}
This section introduces our proposed WARM strategy and RFD module. We firstly explore the factors of max IoU between anchors and each face. Next, anchor matching simulation is performed to evaluate sampling range of face aspect ratio. Then, we demonstrate WARM strategy in detail. Finally, we design RFD module to fit for various aspect ratio of face features.

\subsection{Factors of Face Max IoU}
\label{subsec3.1}
For anchor-based face detection, one of the most important steps is to match ground-truth boxes with well-designed anchors and assign those anchors with positive and negative labels based on their IoUs. However, we discover that sampling positve anchors from each face is not always successful. When anchor design determined, SAM  strategy always fails to sample positive anchors from faces with extreme AR. Therefore, we theoretically explore what affects the max IoU of each face and the reason of sampling failure. \\

\indent{}Formally, the set of anchors is denoted as \emph{A}=\{$a_{i}\}_{i=1}^{m}$, where \emph{i} is the index of anchors and \emph{m} is the number of anchors for all scales. $B_{i}^{a}$ = ($x_{i}^{a}$,$y_{i}^{a}$,$w_{i}^{a}$,$h_{i}^{a}$) is the bounding box of anchor $a_{i}$, where ($x_{i}^{a}$,$y_{i}^{a}$) is the upper-left coordinates and ($w_{i}^{a}$,$h_{i}^{a}$) represents the width and height of this anchor. Besides, \emph{$r^{a}$} is the aspect ratio of all anchors. Similarly, the set of ground-truth faces is denoted as \emph{G}=\{$g_{j}\}_{j=1}^{n}$, where \emph{j} is the index of the ground-truth faces and \emph{n} is the number of these faces. $B_{j}^{g}$ = ($x_{j}^{g}$,$y_{j}^{g}$,$w_{j}^{g}$,$h_{j}^{g}$) is the bounding box of ground-truth face $g_{j}$, where ($x_{j}^{g}$,$y_{j}^{g}$) is the upper-left coordinates and ($w_{j}^{g}$,$h_{j}^{g}$) represents the width and height of this face. In addition, \emph{$r_{j}^{g}$} is the aspect ratio of face \emph{$g_{j}$}. Given a ground-truth face \emph{$g_{j}$}, the max IoU of this face can be computed as in Eq.\ref{eq1}. \\

\begin{equation}
\label{eq1}
\max IoU(g_{j}) = \max_{a_{i} \in A} \frac{B_{j}^{g} \cap B_{i}^{a}}{B_{j}^{g} \cup B_{i}^{a}} = \max_{a_{i} \in A}  \frac{1}{\frac{Area(B_{j}^{g})+Area(B_{i}^{a})}{B_{j}^{g} \cap B_{i}^{a}}-1} ,
\end{equation}

\noindent where $\cap$ and $\cup$ denote the intersection and union of two boxes respectively. Area(.) calculates the area of the bounding box. Furthermore, we recognize that there exists the max value of $B_{j}^{g} \cap B_{i}^{a}$ formulated as in Eq.\ref{eq2}. \\
\begin{equation}
\label{eq2}
\begin{aligned}
\max (B_{j}^{g} \cap B_{i}^{a}) &= \min(w_{j}^{g},w_{i}^{a})*\min(h_{j}^{g},h_{i}^{a}) \\
&= \min(w_{j}^{g},w_{i}^{a})*\min(w_{j}^{g}*r_{j}^{g},w_{i}^{a}*r^{a}) ,
\end{aligned}
\end{equation}

\indent When anchor design determined, the width $w_{j}^{g}$ and AR $r_{j}^{g}$ of ground-truth face $g_{j}$ are the rest factors affecting face maxIoU(\emph{$g_{j}$}). Moreover, anchor setting is well designed in a multi-scale manner. Hence, the AR of faces is the key factor to determine their max IoUs. In other words, different faces have their own max IoU overlap according to AR themselves. Unfortunately, SAM strategy utilizes the same IoU threshold for all faces. As a result, the failure of sampling positive anchors from extreme AR faces is just because the max IoUs of these faces are still lower than sampling IoU threshold. \\
\subsection{Anchor Matching Simulation}
\label{subsec3.2} 
According to the max IoU of faces as discussed in Subsection \ref {subsec3.1}, Anchor Matching Simulation (AMS) is performed on Wider Face training set  to evaluate the sampling range of face aspect ratio. We assume that enought random crop is executed. Thus, each face will have the chance to match anchors with its max IoU. The AMS can be described in the following steps: \\

\begin{table}
	\normalsize
	\setlength{\belowcaptionskip}{5pt}
	\centering 
	\caption{The AMS is performed with different anchor aspect ratio $R^{a}$ and sampling threshold $T_{p}$. The aspect ratio sampling range of matched faces is listed blow. We can approximately describe the sampling range as ARSD.}
	\label{tab1}
	{\begin{tabular}{cccc}
			\toprule
			$T_{p}$ & $R^{a}$ & Range & ARSD  \\
			\midrule
			0.50 & 1.50 & 0.666667 $\sim$ 3.363636 & D(1.50,2.25)  \\
			0.50 & 1.25 & 0.560000 $\sim$ 2.809524 & D(1.25,2.25)  \\
			0.50 & 1.00 & 0.449275 $\sim$ 2.241379 & D(1.00,2.25)  \\
			0.45 & 1.00 & 0.388889 $\sim$ 2.586207 & D(1.00,2.59)  \\
			0.40 & 1.00 & 0.333333 $\sim$ 3.055556 & D(1.00,3.06)  \\
			0.35 & 1.00 & 0.285714 $\sim$ 3.666667 & D(1.00,3.67)  \\
			\bottomrule
		\end{tabular}}
	\end{table}	

\begin{itemize}
\item Step 1: Construct a high-recall anchor design. \\
\item Step 2: Calculate max IoU of each face. \\
\item Step 3: Judge if current face can match positive anchors. \\
\item Step 4: Record the sampling range of face aspect ratio. \\
\end{itemize}

\indent{} To be more specific, we firstly construct a high-recall anchor design. The anchor size ranges from 4 to 512 pixels while the anchor stride is $\sqrt{2}$. All anchors have the same AR. Next, the max IoU of each face can be calculated as shown in E.q\ref{eq1} and E.q\ref{eq2}. Then, we compare face max IoU with positive sampling threshold $T_{p}$. If the max IoU of current face is greater than $T_{p}$, the positive anchors related to this face can be added into training samples. Finally, we update the sampling range of face AR. \\

\indent{} The result of AMS is listed in Tab. \ref{tab1}. Here we take SAM strategy as an example. From the first three rows, we can see that sampling range centers around the AR of anchors. More experimental results show that anchor AR of 1.0 can achieve higher performance. Besides, we find that the sampling range enlarges gradually as positive sampling threshold reduces as listed in Tab. \ref{tab1} of last four rows. For convenience, we define Aspect Ratio Sampling Domain (ARSD) to approximately describe the sampling range of face AR as follow: \\

\begin{myDef}
	(Aspect Ratio Sampling Domain). Given an anchor design A, where $r^{a}$ is the aspect ratio of anchors. M denotes the anchor matching strategy. For a ground-truth face set G, the aspect ratio sampling domain D($r^{a}$,$\eta$) is difined as
	\begin{equation}
	D(r^{a},\eta) = \{x|r^{a}/\eta<x<r^{a}\eta\},
	\label{eq3}
	\end{equation}
\end{myDef}
where $\eta$ is the radius of sampling domain. Thus, the left and right ARSD can be descirbed as follows: \\	
\begin{equation}
D^{-}(r^{a},\eta) = \{x|r^{a}/\eta<x<r^{a}\},
\label{eq4}
\end{equation}
	
\begin{equation}
D^{+}(r^{a},\eta) = \{x|r^{a} \leq x<r^{a}\eta\},
\label{eq5}
\end{equation}

\noindent{} When positive sampling threshold \emph{$T_{p}$} of SAM is set to 0.5, we notice that the ARSD is \emph{D}(1.00,2.25) as listed in third row of Tab. \ref{tab1}. However, statistics show that the AR of more than 99.96\% faces on the Wider Face training set is in \emph{D}(1.00,5.00). Therefore, a considerable part of extreme AR faces is neglected during anchor matching phase as seen in Fig. \ref{fig1}.  
\subsection{Wide Aspect Ratio Matching}
\label{subsec3.3}
Current anchor matching strategy usually consists of two steps: SAM and anchor compensation. Each face firstly attemps to match all anchors with IoU higher than predefined threshold. However, some of these ground-truth faces may be unmatched in this step, especially for extreme AR faces. Then, unmatched faces would be compensated with the rest anchors that have highest IoU with them in current iteration. Obviously, compensated anchors may reduce the detection performance since these anchors have lower IoU with unmatched faces. Therefore, we believe that current anchor matching strategy is neither flexible nor sufficient to match the anchors in face detection.

To address this issue, we propose a Wide Aspect Ratio Matching strategy to collect more representative positive anchors from a wide range of face aspect ratio. The core idea is to construct variable positive threshold for extreme AR faces. We firstly determine the sampling domain of extreme AR faces according to the result of AMS as follows.

\begin{equation}
E(\eta_{1}, \eta_{0})=D(r^{a},\eta_{1}) \setminus D(r^{a},\eta_{0})
\label{eq6}
\end{equation}

\begin{figure}[htbp]
	\centering
	\includegraphics[width=4in,height=2in]{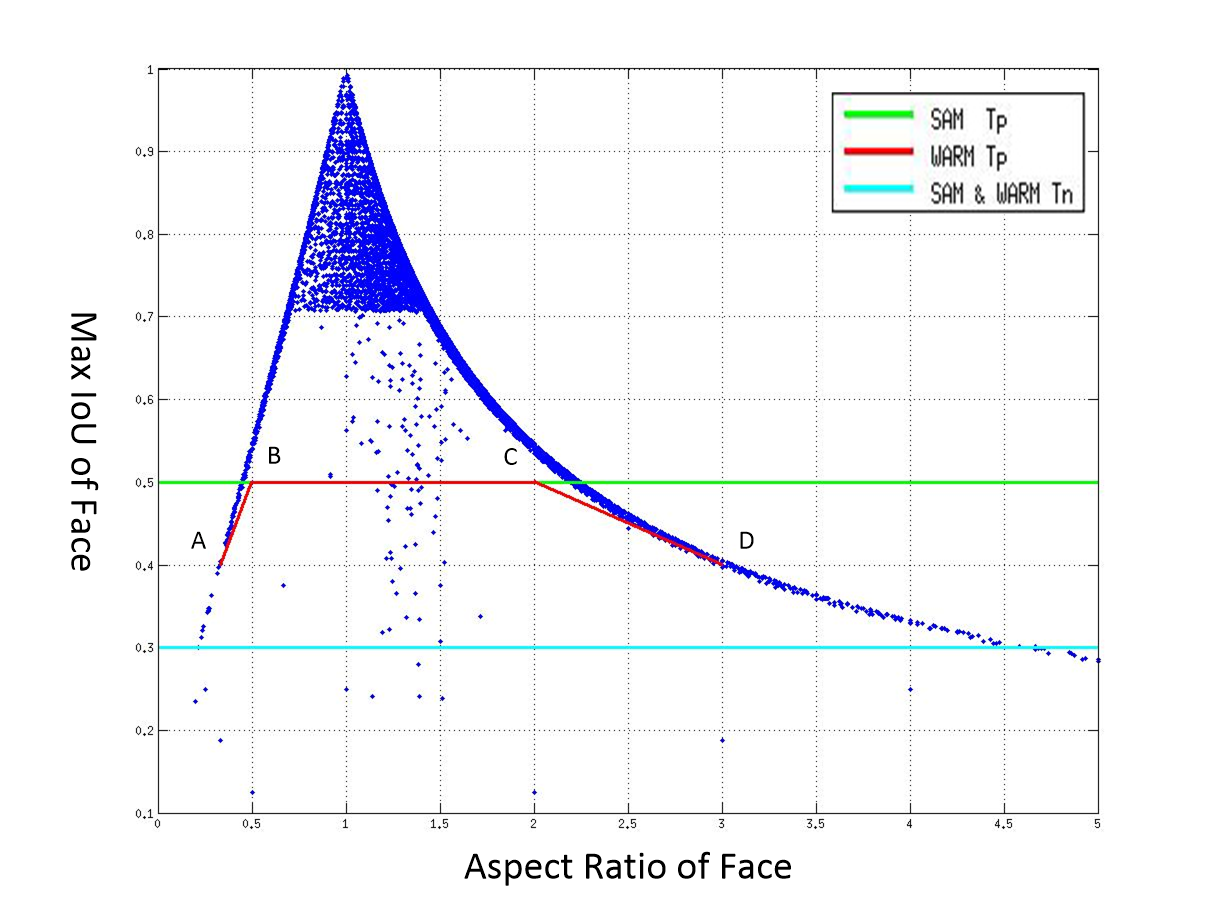}
	\caption{The distribution of face aspect ratio and their max IoUs. The coordinate of blue dot represents the aspect ratio of each face and their max IoUs. The green line is the positive threshold boundary of SAM while the red line is our proposed WARM. Both of them adopt the cyan line as negative threshold boundary.}
	\label{fig2}
\end{figure}

\noindent where $D(r^{a},\eta_{1})$ represents the total sampling domain in our WARM strategy and $D(r^{a},\eta_{0})$ is a subset of this sampling domain. According to the result of AMS, $\eta_{0}$ is set to 2.0. After that, the difference set $E(\eta_{1}, \eta_{0})$ of these two ARSDs is the sampling domain of extreme AR faces. Furthermore, the left and right sampling domain of extreme AR faces can be noted below.

\begin{equation}
E^{-}(\eta_{1}, \eta_{0})=D^{-}(r^{a},\eta_{1}) \setminus D^{-}(r^{a},\eta_{0})
\label{eq7}
\end{equation}
\begin{equation}
E^{+}(\eta_{1}, \eta_{0})=D^{+}(r^{a},\eta_{1}) \setminus D^{+}(r^{a},\eta_{0})
\label{eq8}
\end{equation}

\noindent Then, we construct a variable positive threshold function in the sampling domain of extreme AR faces while follow the SAM strategy in the rest sampling domain. For simplicity, the linear function is applied. The positive threshold of our WARM can be formulated as follow:

\begin{equation}
T_{p}=\left\{
\begin{aligned}
T_{0} - \delta * \theta(r^{g}_{j}),	&	&{\ r^{g}_{j} \in E(\eta_{1}, \eta_{0})} \\
T_{0},	& &{otherwise}
\end{aligned}
\right. 
\label{eq9}
\end{equation}
\noindent where \emph{$T_{0}$} is the initial value of positive threshold and $\delta$ represents the amplitude of positive threshold. Similar to SAM, \emph{$T_{0}$} is set to 0.5. Besides, $\theta(r^{g}_{j})$ reflects the change rate of positive threshold associated with the AR $r^{g}_{j}$ of each face. When AR of faces far away from current anchor AR, positive threshold of these face should decrease gradually. Thus, a simple implementation of $\theta(x)$ is given below.
\begin{equation}
\theta(x) = \left\{
\begin{aligned}
\frac{\max(x)-x}{\max(x)-\min(x)},	&	&{\ x \in E^{-}(\eta_{1}, \eta_{0})} \\
\frac{x-\min(x)}{\max(x)-\min(x)} ,	& &{\ x \in E^{+}(\eta_{1}, \eta_{0})}
\end{aligned}
\right. 
\label{eq10}
\end{equation}

\indent{} To visualize our proposed WARM strategy, we plot the scatter diagram of all training faces in Fig. \ref{fig2}. The coordinate of each blue dot represents the AR of a face and its maximum IoU. It should be noted that face max IoU is the IoU value of the best matching anchor with this face. The green line is the positive threshold boundary of SAM while the red line is our proposed WARM. Specifically, line AB and CD represent the variable positive threshold boundary of extreme AR faces. Instead of anchor compensation, variable positive threshold can match extreme AR faces with higher IoU anchors. Similarly, both of them adopt the cyan line as negative threshold boundary. When $\delta$ is set to 0, our proposed WARM degenerates to SAM strategy. Therefore, our proposed WARM method could be a general strategy for anchor-based single-stage face detection.

\subsection{Receptive Field Diversity}
\label{subsec3.4}
\begin{figure}
	\centering
	\includegraphics[width=3in,height=2in]{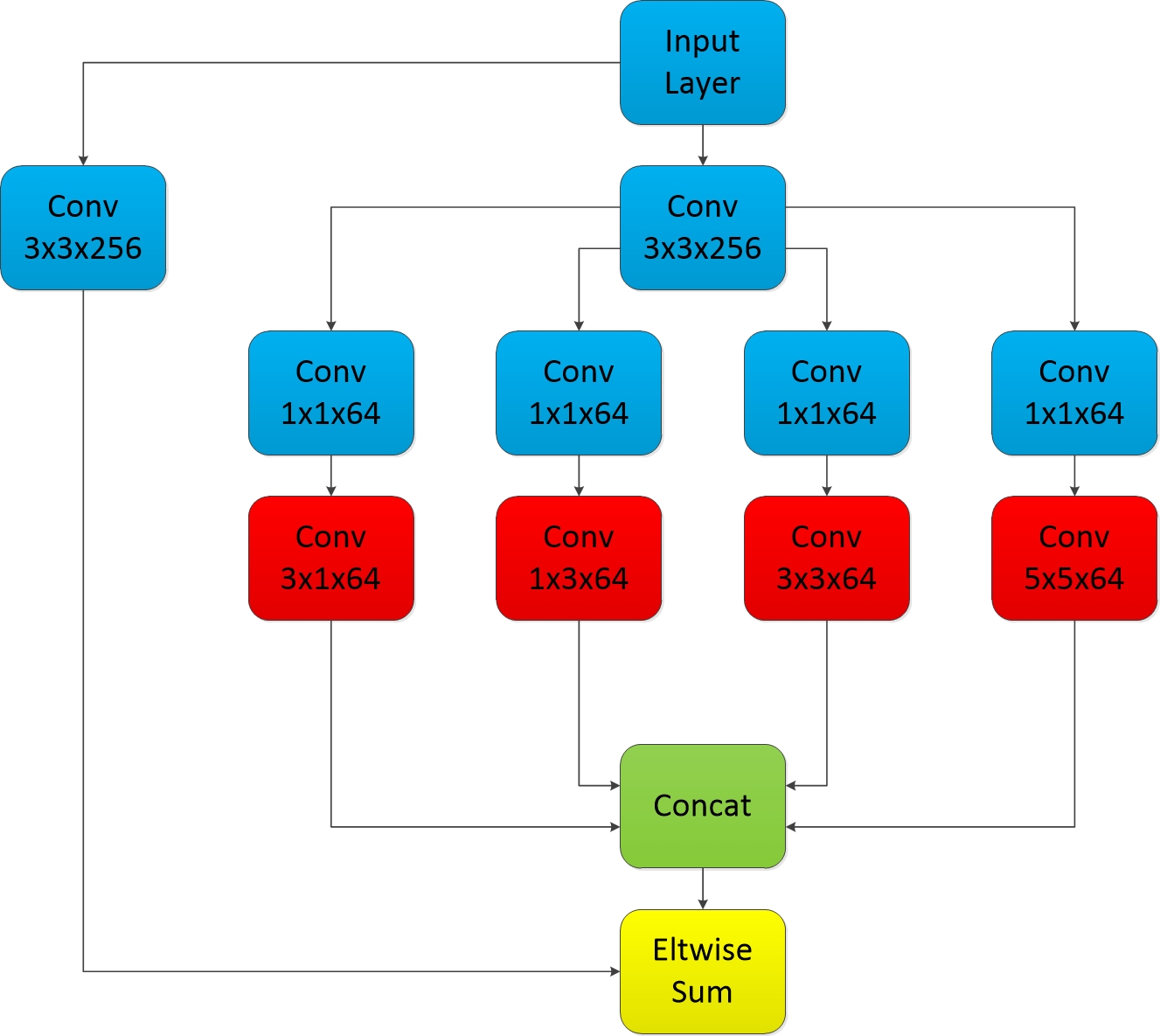}
	\caption{The structure of Receptive Field Diversity module.}
	\label{fig3}
\end{figure}
Current feature enhancement modules usually enlarge receptive field by mean of symmetric convolution kernels. The singleness of the receptive field is no longer fit for extreme AR faces. To address this issue, we propose a novel feature enhancement module, named Receptive Field Diversity (RFD) module, to provide more diverse aspect ratio and large receptive field simultaneously. Both symmetric and asymmetric convolution kernels \cite{ACNet} are used to adapt the diversity of face AR.

\indent{} Fig. \ref{fig3} illustrates the structure of RFD module, which is inspired by Inception \cite{GoogleNet} and ResNet blocks. The RFD module adopts a 4-path structure. In particular, we firstly utilize a 1x1 convolution layers to reduce the channel number to one quarter of the input layer. Then, the 3$\times$1, 1$\times$3, 3$\times$3 and 5$\times$5 convolution kernels are employed to provide diversity receptive field. Finally, these 4-path feature maps are concatenated togather. Besides, we apply a shortcut path, which maintain the original receptive field, to sum up with concatenated features above.  

\section{Experiments}
\label{sec4}

\subsection{Experimental Setup}
\label{sec4.1}
In this section, we introduce the backbone, anchor design, data agumentation, loss function and other implementation details.\\
\indent{} \textbf{Backbone.} We adopt ResNet-50 with 5-level feature pyramid structure as the backbone network in our method. The feature maps are extracted from those four residual blocks, denoted as C2, C3, C4 and C5. P2, P3, P4, and P5 are the fused feature maps \cite{FPN} corresponding to C2, C3, C4 and C5, while P6 is just down-sampled by two 3x3 convolution layers after C5.\\
\indent{} \textbf{Anchor Design.} For anchor generation, we assign  \{4, 4$\sqrt{2}$, 8\} in \textit{P$_{2}$}, \{8$\sqrt{2}$, 16, 16$\sqrt{2}$\} in \textit{P$_{3}$}, \{32, 32$\sqrt{2}$, 64\} in \textit{P$_{4}$}, \{64$\sqrt{2}$, 128, 128$\sqrt{2}$\} in \textit{P$_{5}$}, and \{256, 256$\sqrt{2}$, 512\} in \textit{P$_{6}$}. All anchors have aspect ratio of 1.0.\\
\indent{} \textbf{Data Augmentation.} We randomly crop \cite{Pyramidbox,S3FD} square patches from the original images and resize these patches into 640$\times$640. Except for random crop, we also utilize random horizontal flip with probability of 0.5 and photo-metric color distortion \cite{S3FD} to augment training data.\\
\indent{} \textbf{Loss Function.} We apply the multi-task loss as our objective function. Specifically, Focal loss \cite{FocalLoss} is used for the binary classification while Smooth-L1 loss is for the bounding box regression.\\
\indent{} \textbf{Optimization details.} We use stochastic gradient descent (SGD) with momentum 0.9 and weight decay $5\times10^{-5}$ to fine-tune our detection models. The learning rate is set to 0.01 for the first 60 epochs, and decreases to $10^{-3}$ and $10^{-4}$ for the next two 30 epochs. Besides, OHEM \cite{OHEM} is applied to alleviate significant imbalance between the positive and negative training examples with a ratio of 1:3. During training phase, 256 detections per module are selected for each image. During inference phase, each module outputs 1000 detection results whose confidence scores are all higher than the threshold of 0.02. Finally, we perform NMS with a threshold of 0.3 on the outputs of all modules together. Our method is implemented in MXNet \cite{Mxnet} and all the experiments are trained on 2 NVIDIA GeForce GTX 1080Ti GPUs in parallel.\\

\subsection{Datasets}
\label{sec4.2}
WIDER FACE dataset: It consists of 32,203 images with 393,703 labeled face boxes with a high degree of variability in scale, pose and occlusion. These images are split into training (40\%), validation (10\%), and testing (50\%) sets by randomly sampling from 61 event classes. Faces in this dataset are classified into Easy, Medium, and Hard subsets according to their detection difficulty. We train all models on the training set of the WIDER FACE dataset while evaluate on its validation and test sets. Ablation studies are also performed on the validation set.\\

FDDB dataset: It contains 2845 images and 5171 annotated faces. Most of these faces have large scale, high resolutions or slightly occlusion sometimes. Different from WIDER FACE, faces in the FDDB dataset are labeled by bounding ellipses. In order to verify generalization ability of our method, we perform the evaluation on the FDDB dataset.

\subsection{Ablation Study}
\label{sec4.3}
In this subsection, we conduct ablation studies to evaluate the effectiveness of our proposed WARM and RFE. For fair comparisons, we use the same settings as described in Subsection \ref{sec4.1} for all the experiments.
\begin{table*}
	\normalsize
	\setlength{\belowcaptionskip}{5pt}
	\centering
	\caption{Varying $\eta_{1}$, $\delta$ for WARM on WIDER FACE validation set.}
	\begin{tabular}{cccccc}
		\toprule
		\multirow{2}{*}{Method} &
		\multirow{2}{*}{$\eta_{1}$} & \multirow{2}{*}{$\delta$} & \multicolumn{3}{c}{AP} \\
		\cline{4-6}
		& & & Easy & Medium & Hard \\
		\midrule
		SAM & 2.25 & 0.00 & 0.959 & 0.950 & 0.898 \\
		\midrule
		& 2.50 & 0.05 & 0.962 & 0.952 & 0.899 \\
		& 2.50 & 0.10 & 0.961 & 0.952 & 0.901 \\
		WARM & 2.50 & 0.15 & 0.958 & 0.949 & 0.897 \\
		& 3.00 & 0.10 & \textbf{0.962} & \textbf{0.953} & \textbf{0.902} \\
		& 4.00 & 0.10 & 0.960 & 0.951 & 0.898 \\
		\bottomrule
	\end{tabular}
	\label{tab2}
\end{table*}

\indent{} \textbf{The effect of Wide Aspect Ratio Matching strategy.} We discuss the effect of two hyper-parameters in our proposed WARM strategy. The performance under different $\eta_{1}$, $\delta$ (defined in Subsection \ref{subsec3.3}) is shown in Table \ref{tab2}. Compared with SAM, our proposed WARM can collect more positive anchors from extreme AR faces, whose ARSD is in E($\eta_{1}$,2.0). Besides, these extra collected positive anchors have higher IoU, ranging from 0.5-$\delta$ to 0.5, related to their matched extreme AR faces. After multiple ablative experiments, we find the optimal hyper-parameters. When $\eta_{1}$ and $\delta$ are set to 3.0 and 0.1, our proposed WARM can increase the detection performance of 0.3\%(Easy), 0.3\%(Medium), and 0.4\%(Hard) separately.\\

\begin{table*}
	\normalsize
	\setlength{\belowcaptionskip}{5pt}
	\centering
	\caption{Effectiveness of RFD module on the AP performance.}
	\begin{tabular}{cccc}
		\toprule
		\multirow{2}{*}{Component} &
		\multicolumn{3}{c}{AP} \\
		\cline{2-4}
		& Easy & Medium & Hard \\
		\midrule
		SSH & 0.959 & 0.950 & 0.898 \\
		RFD & 0.961 & 0.951 & 0.900 \\
		\midrule
		Ours(WARM + RFD) & \textbf{0.965} & \textbf{0.955} & \textbf{0.904} \\
		\bottomrule
	\end{tabular}
	\label{tab3}
\end{table*}

\indent{} \textbf{The effect of Receptive Field Diversity module.} To demonstrate the effectiveness of RFD module, we conduct the comparation experiment between SSH and RFD module as shown in Tab. \ref{tab3}. Previous feature enhancement modules utilize square convolutional kernels to enlarge the receptive field. Here, we take SSH detection module as an example. In order to adapt the diversity of face AR, both the symmetric and asymmetric convolution kernels are applied  in our proposed RFD module.  From the Tab. \ref{tab3}, we can see that RFD module can further enhance the feature maps and improve the detection performance of 0.2\%(Easy), 0.1\%(Medium), and 0.2\%(Hard) respectively.\\

\indent{} Combining the WARM strategy and RFD module, our method achieves a promising detection performance as shown in the last row of Tab. \ref{tab3}.
\subsection{Evaluation on Benchmark}
\label{sec4.4}
We evaluate our proposed method against state-of-the-art methods on two public face detection benchmarks.
\subsubsection{Wider Face Dataset}
\label{sec4.4.1}

\begin{figure*}
	\begin{minipage}{0.3\linewidth}
		\centerline{\includegraphics[width=1.4in]{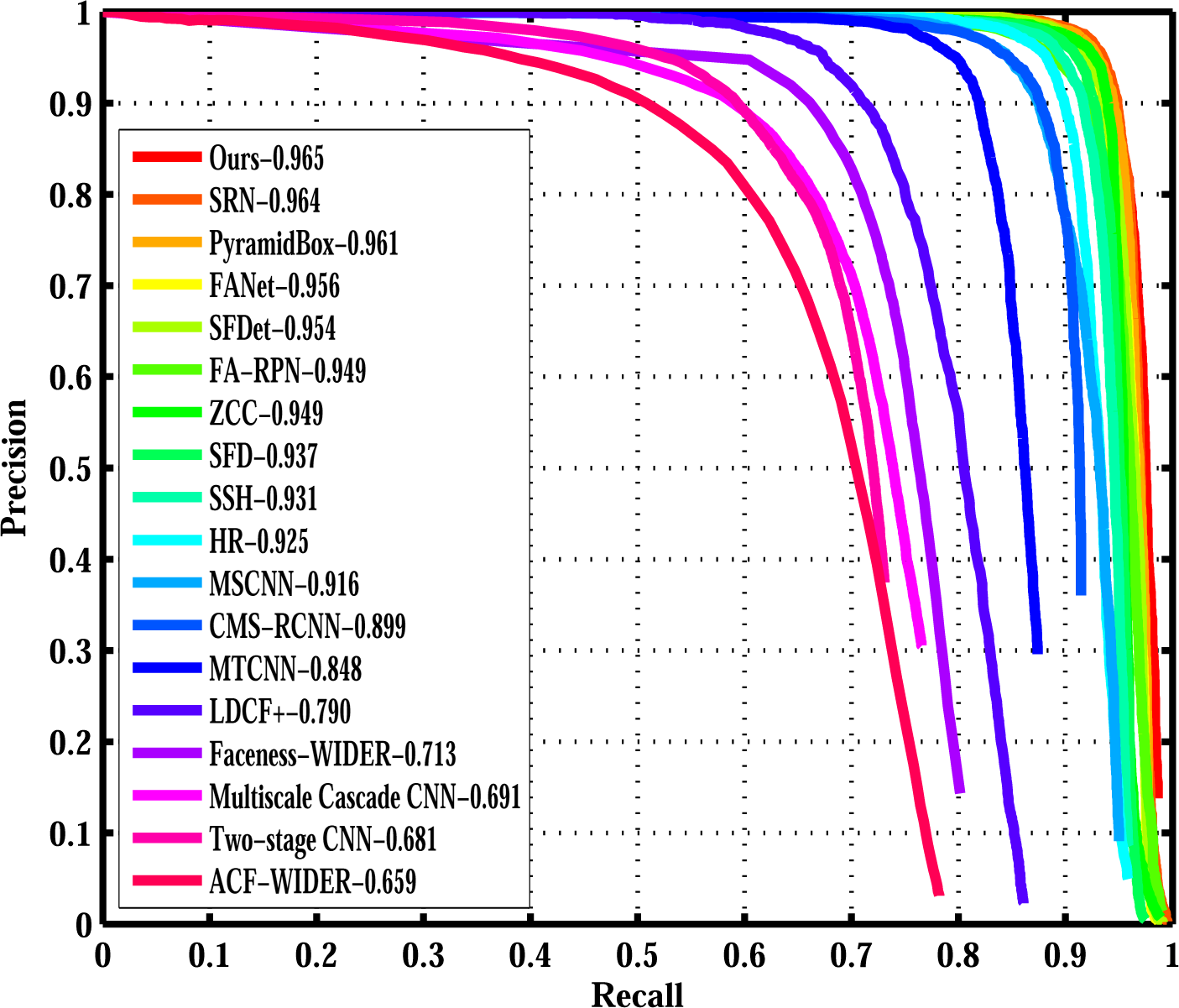}}
		\centerline{(a)	Val: Easy}
	\end{minipage}
	\hfill
	\begin{minipage}{0.3\linewidth}
		\centerline{\includegraphics[width=1.4in]{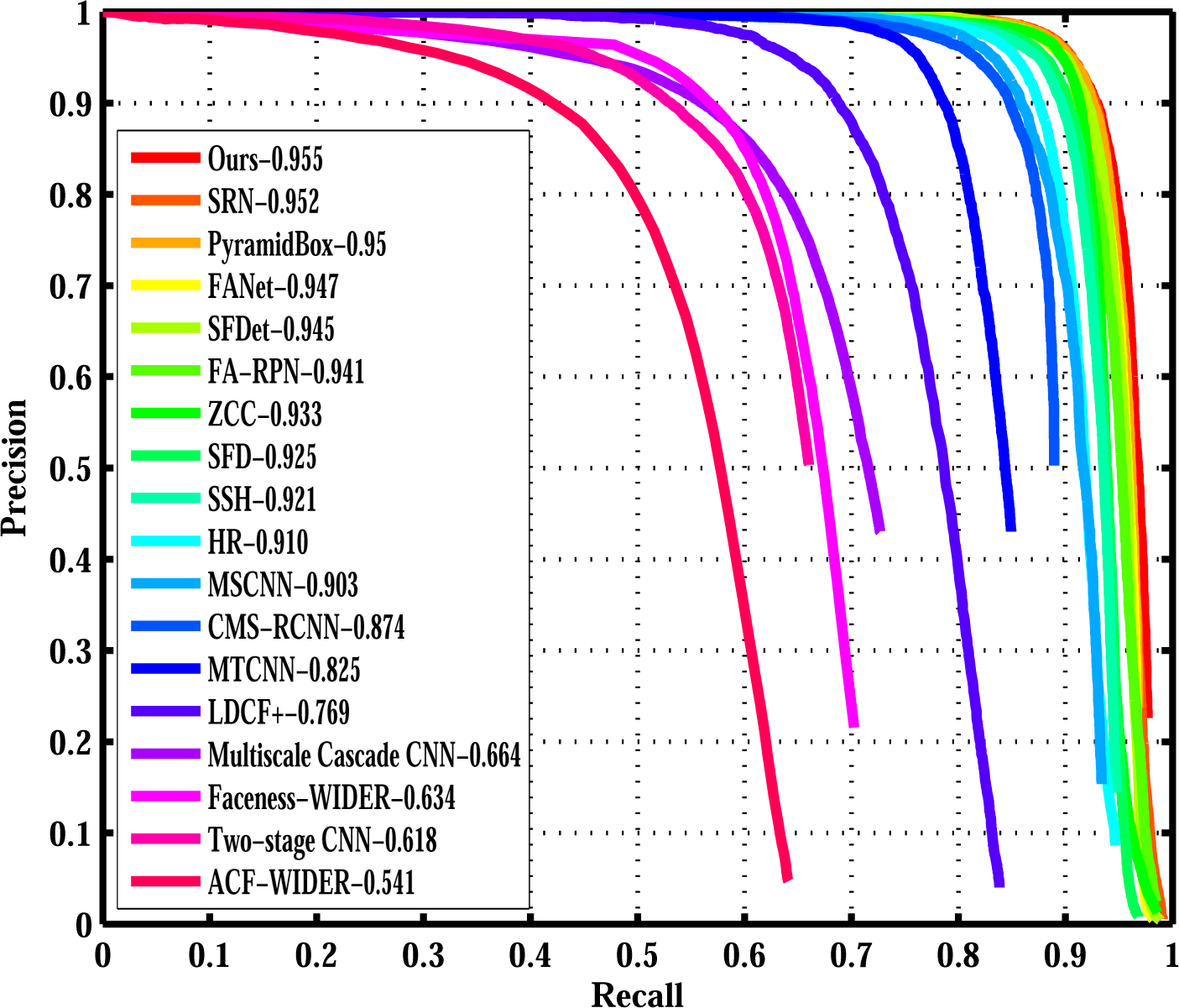}}
		\centerline{(b)	Val: Medium}
	\end{minipage}
	\hfill
	\begin{minipage}{0.3\linewidth}
		\centerline{\includegraphics[width=1.4in]{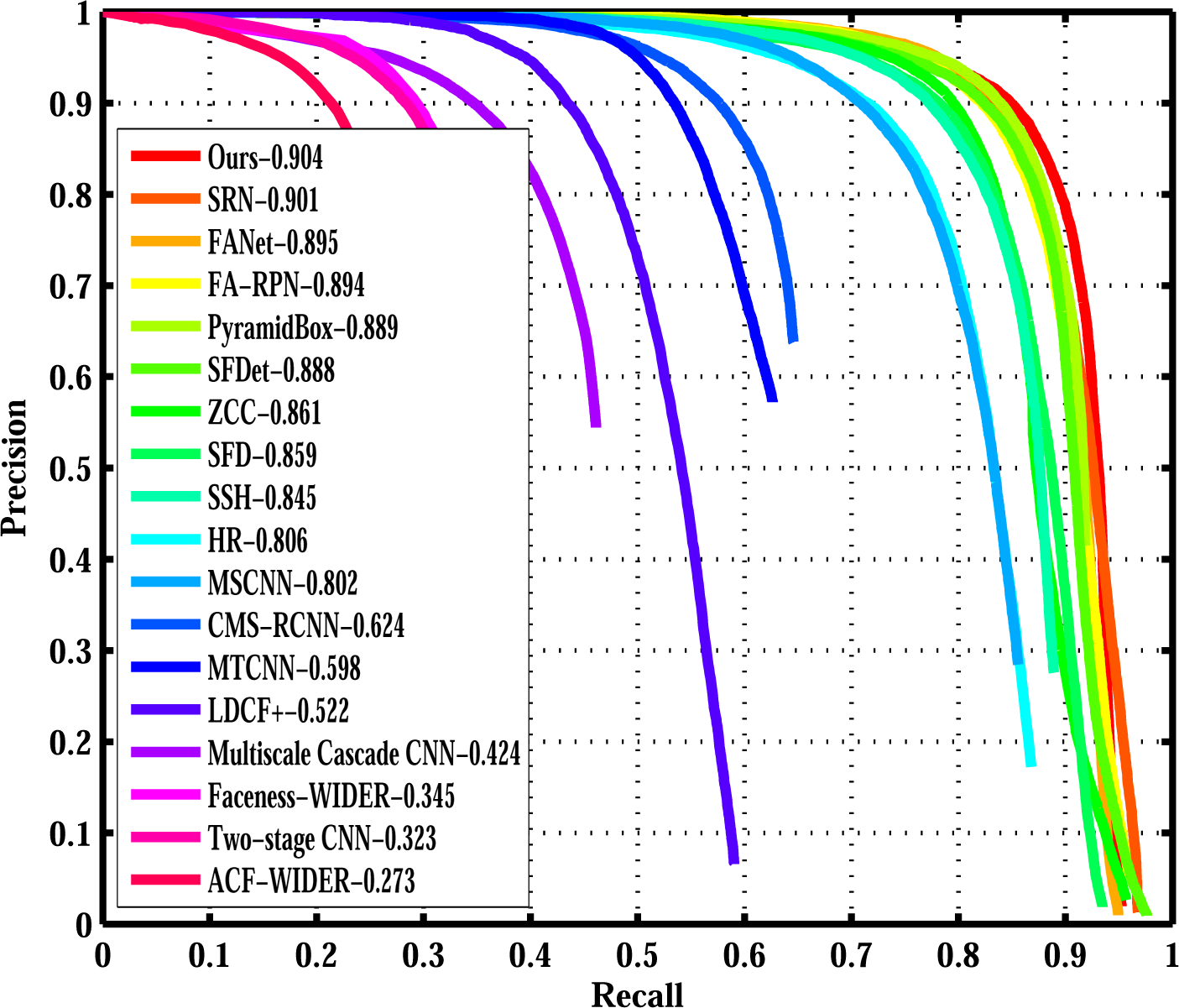}}
		\centerline{(c)	Val: Hard}
	\end{minipage}
	
	\vfill
	\begin{minipage}{0.3\linewidth}
		\centerline{\includegraphics[width=1.4in]{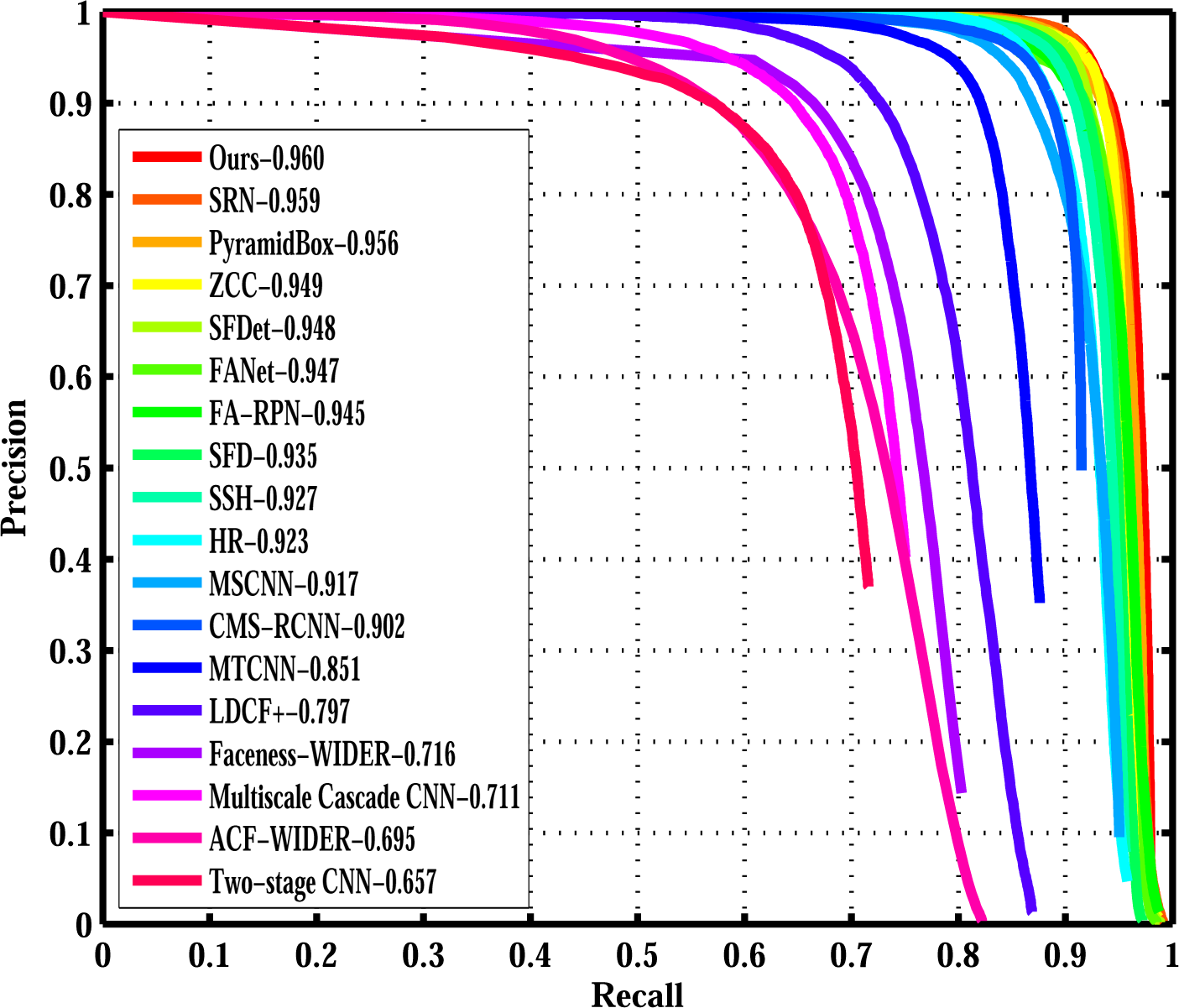}}
		\centerline{(d) Test: Easy}
	\end{minipage}
	\hfill	
	\begin{minipage}{0.3\linewidth}
		\centerline{\includegraphics[width=1.4in]{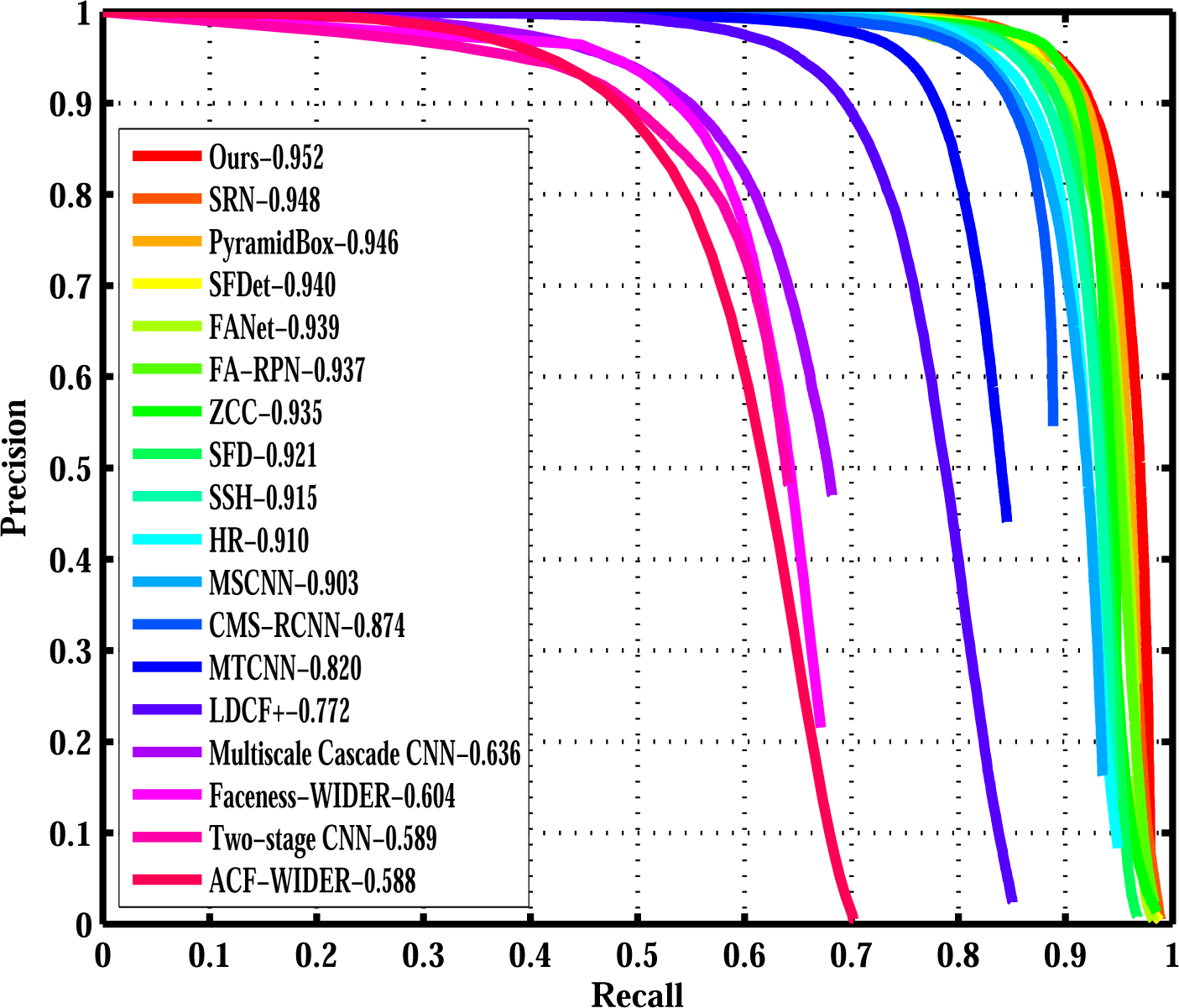}}
		\centerline{(e) Test: Medium}
	\end{minipage}
	\hfill
	\begin{minipage}{0.3\linewidth}
		\centerline{\includegraphics[width=1.4in]{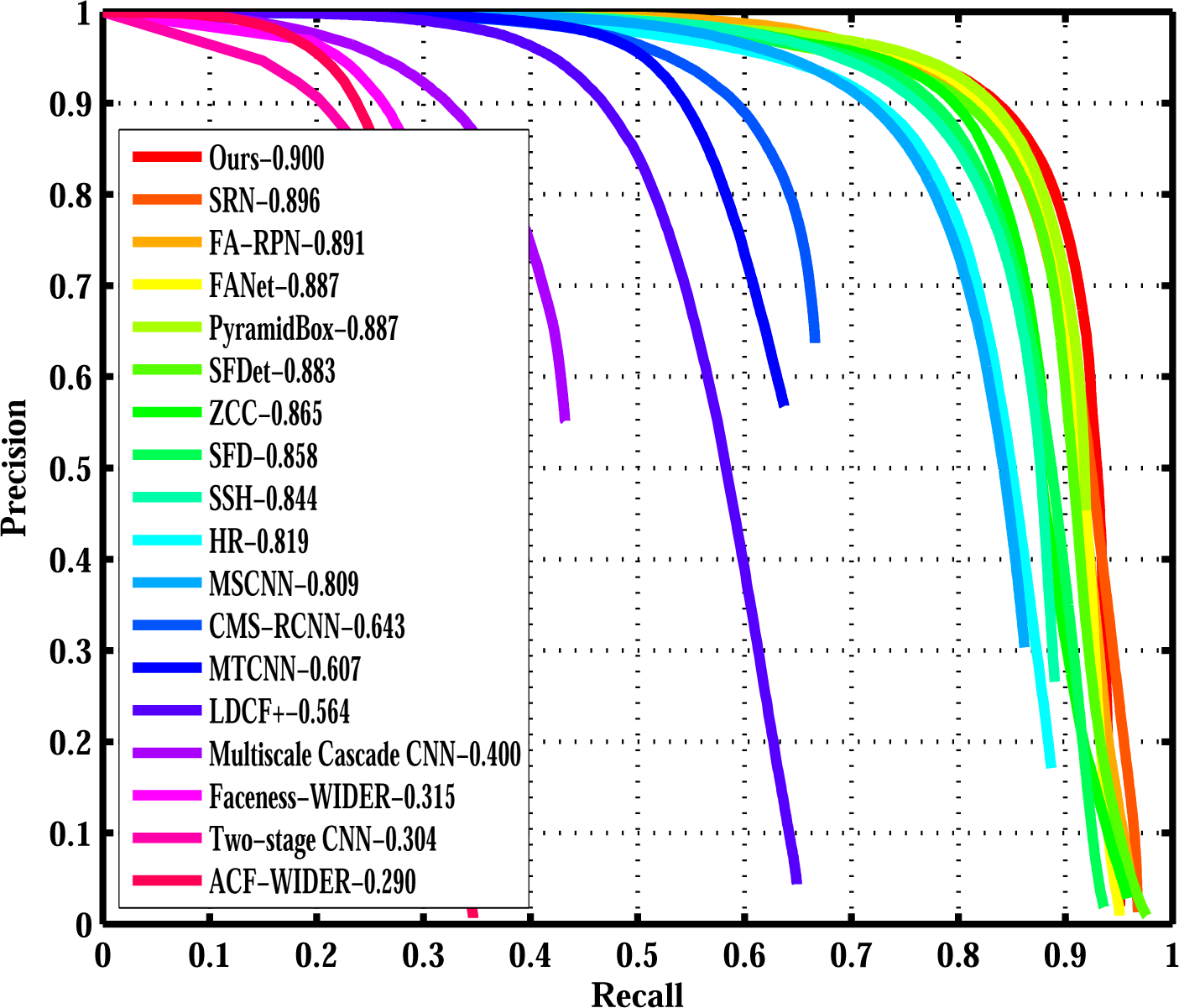}}
		\centerline{(f) Test: Hard}
	\end{minipage}
	
	\caption{Precision-recall curves on WIDER FACE validation and test sets.}
	\label{fig:4}
\end{figure*}

Our model is trained on the training set and evaluate on its validation and testing set against the recently published state-of-the-art face detection methods including SRN \cite{SRN}, PyramidBox \cite{Pyramidbox}, FANet \cite{FANet}, SFDet \cite{SFDet}, FA-RPN \cite{FA-RPN}, ZCC \cite{Zhu}, S$^{3}$FD \cite{S3FD}, SSH \cite{SSH}, HR \cite{HR}, MSCNN \cite{MSCNN}, CMS-RCNN \cite{CMS-RCNN}, MTCNN \cite{MTCNN}, LDCF+ \cite{LDCF+}, Faceness \cite{Faceness}, Multiscale Cascade CNN \cite{WIDERFACE}, ACF \cite{ACF} and Two-stage CNN \cite{WIDERFACE}. The precision-recall curves and AP values on WIDER FACE validation and testing sets are shown in Fig. \ref{fig:4}. It can be seen that our proposed method consistently achieves the best performance across all the three subsets. It achieves the promising average precision in all level faces, i.e. 0.965 (Easy), 0.955 (Medium), and 0.904 (Hard) for validation set, and 0.960 (Easy), 0.952 (Medium), and 0.900 (Hard) for testing set. The result in Fig. \ref{fig:4} demonstrates the effectiveness of our proposed method.

\subsubsection{FDDB Dataset}
\label{sec4.4.2}
\begin{figure*}
	\centering
	\begin{minipage}{0.45\linewidth}
		\centerline{\includegraphics[width=2.0in,height=1.8in]{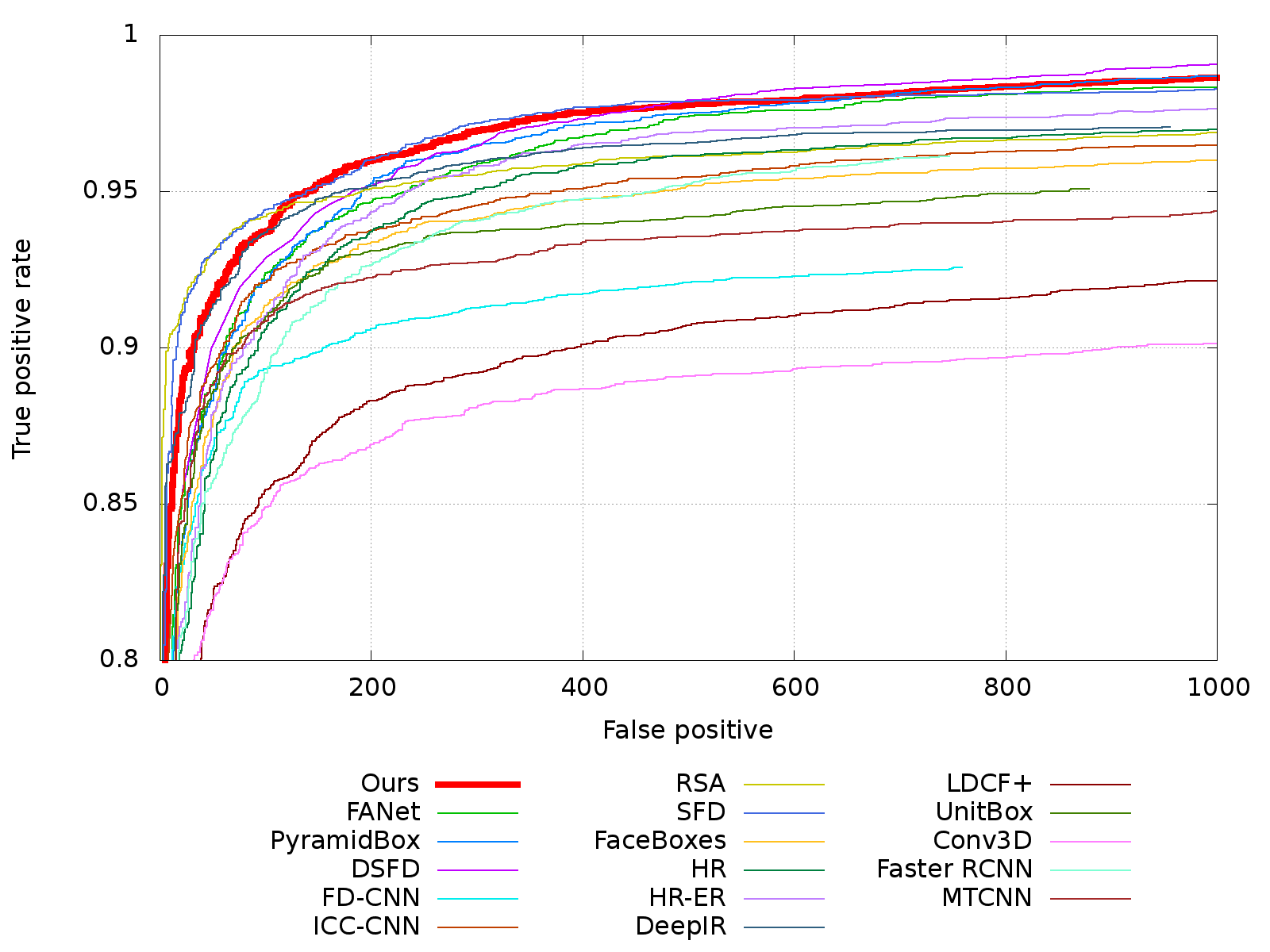}}
		\centerline{(a) Discontinuous ROC curves}
	\end{minipage}
	\hfill
	\begin{minipage}{0.45\linewidth}
		\centerline{\includegraphics[width=2.0in,height=1.8in]{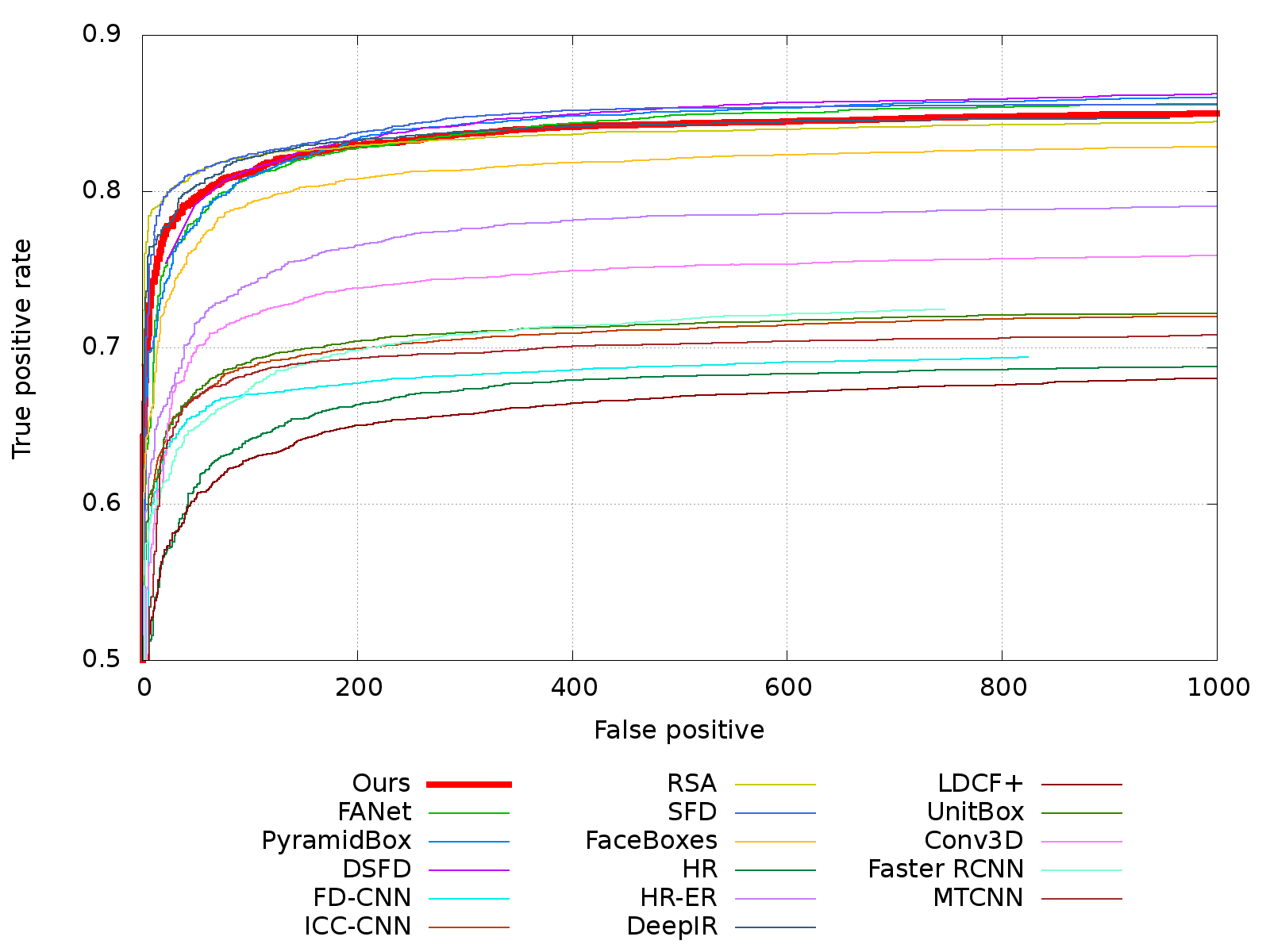}}
		\centerline{(b) Continuous ROC curves}
	\end{minipage}
	\caption{Evaluation on the FDDB dataset.}
	\label{fig:5}
\end{figure*}

We directly use the same detection model above to perform the evaluation on FDDB dataset. Specifically, the shortest side of the input images is set to 400 pixels while the larger side is  less than 800 pixels. We compare our method against the recently published state-of-the-art methods including FANet \cite{FANet}, PyramidBox \cite{Pyramidbox}, DSFD \cite{DSFD}, FD-CNN \cite{FD-CNN}, ICC-CNN \cite{ICC-CNN}, RSA \cite{RSA}, S$^{3}$FD \cite{S3FD}, FaceBoxes \cite{FaceBoxes}, HR \cite{HR}, HR-ER \cite{HR}, DeepIR \cite{DeepIR}, LDCF+ \cite{LDCF+}, UnitBox \cite{UnitBox}, Conv3D \cite{Conv3D}, Faster RCNN \cite{FaceFasterRCNN} and MTCNN \cite{MTCNN} on FDDB dataset. For a more fair comparison, the predicted bounding boxes are converted to bounding ellipses. Fig. \ref{fig:5} shows the discrete ROC curves and continuous ROC curves of these methods on the FDDB dataset respectively. As can be seen, our proposed method consistently achieves a relatively higher performance in terms of both the discrete ROC curves and continuous ROC curves. These results demonstrate the effectiveness and good generalization capability of our proposed method.

\subsection{Qualitative Results}
\label{sec4.6}

\begin{figure*}
	\begin{minipage}{\linewidth}
		\centerline{
				\includegraphics[width=1.1in]{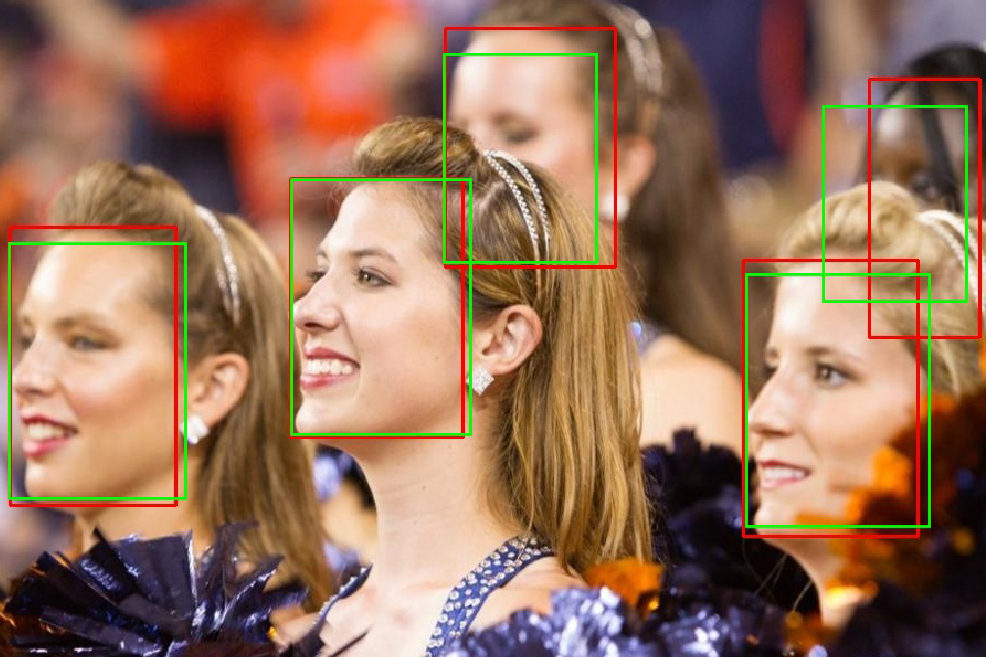}
				\includegraphics[width=1.1in]{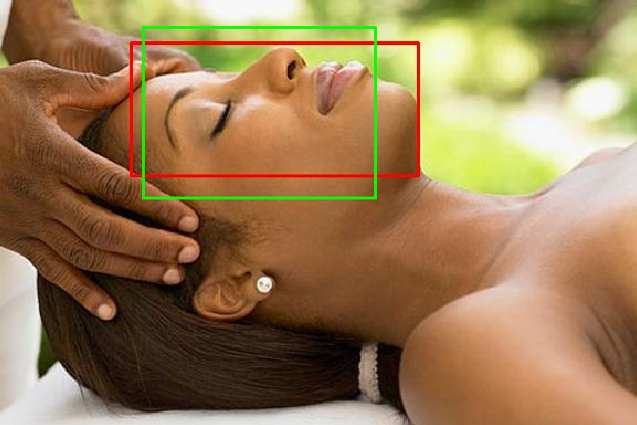}
				\includegraphics[width=1.1in]{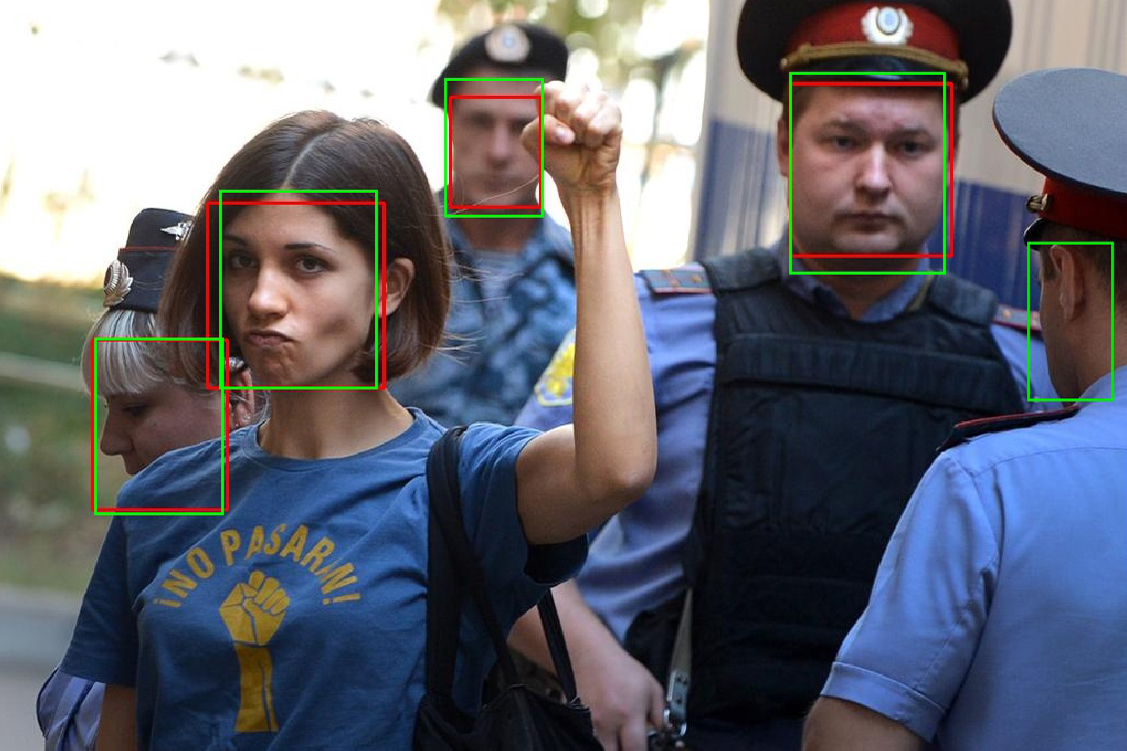}
				\includegraphics[width=1.1in]{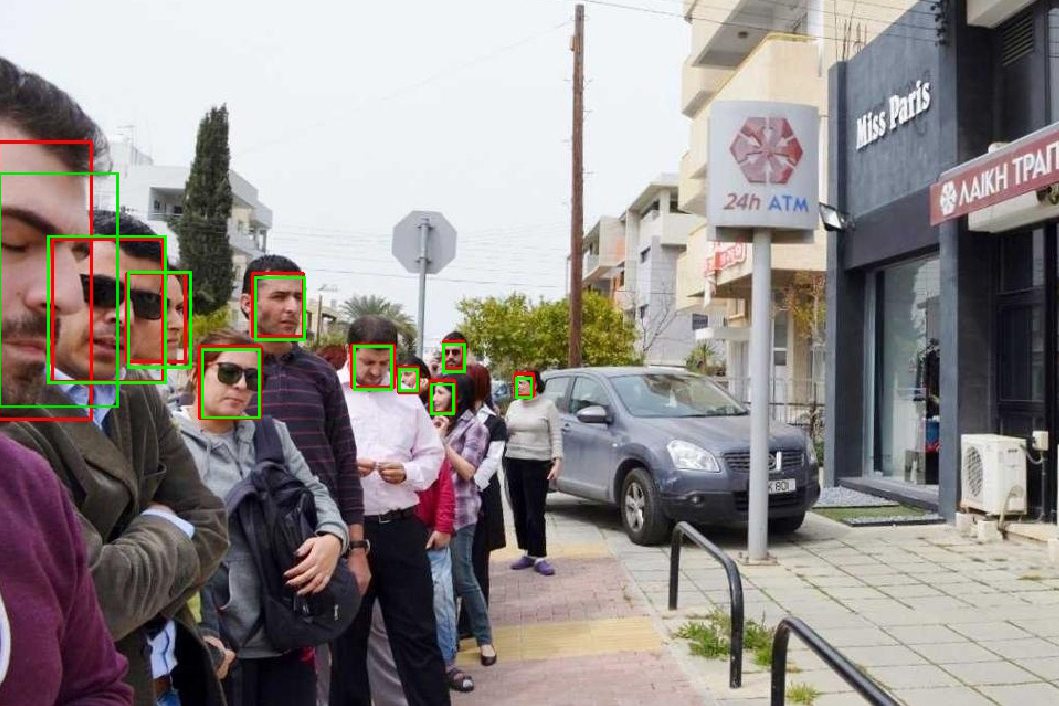}
		}
		\centerline{
				\includegraphics[width=1.1in]{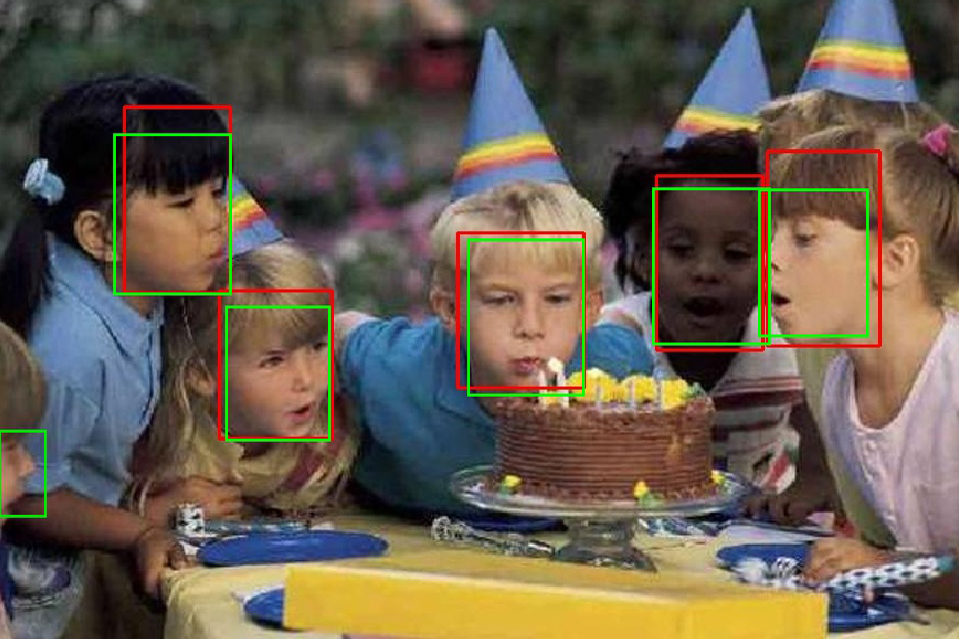}
				\includegraphics[width=1.1in]{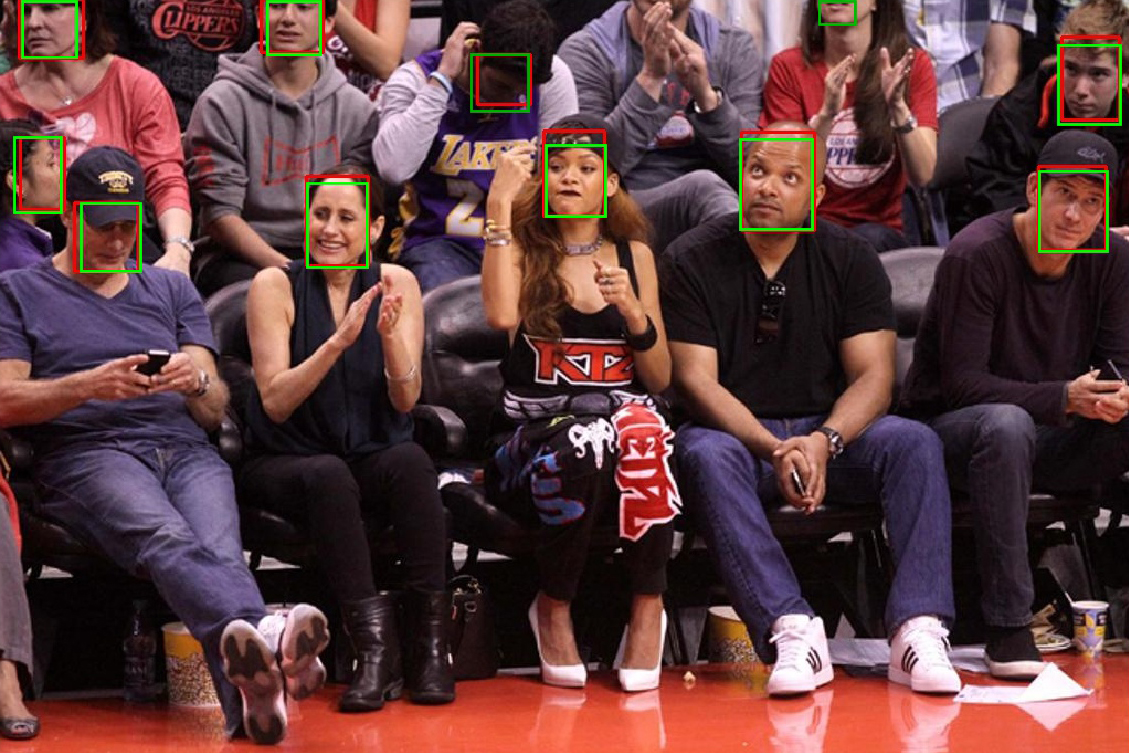}
				\includegraphics[width=1.1in]{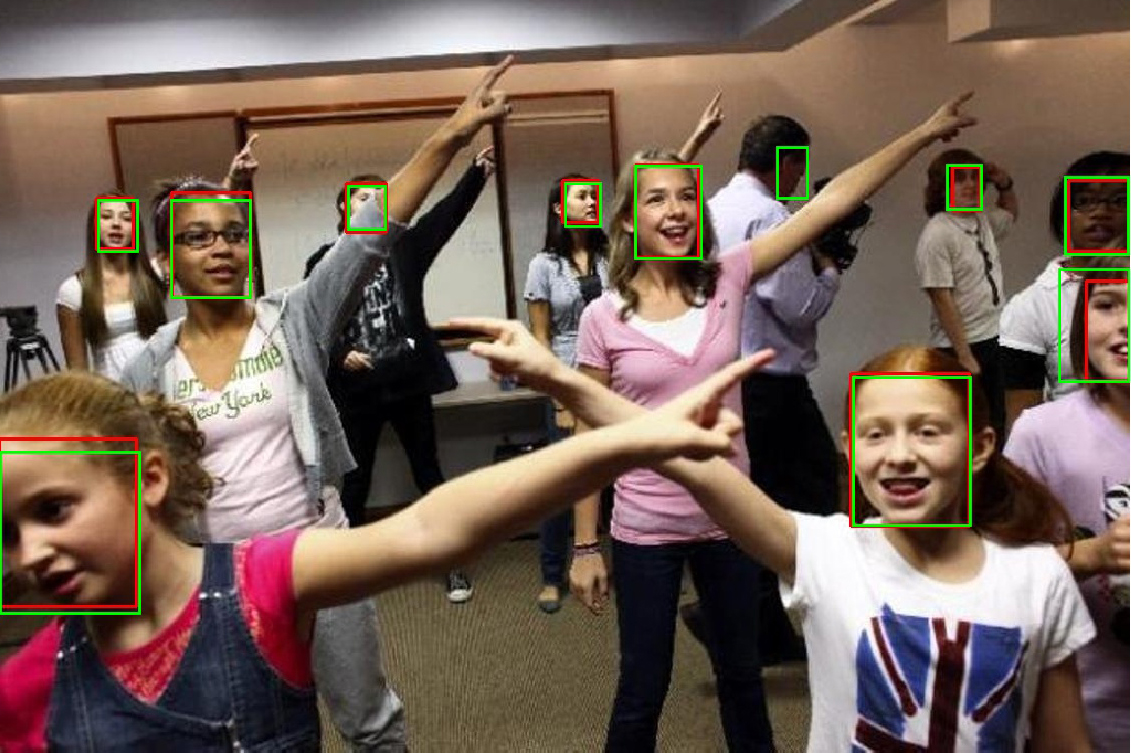}
				\includegraphics[width=1.1in]{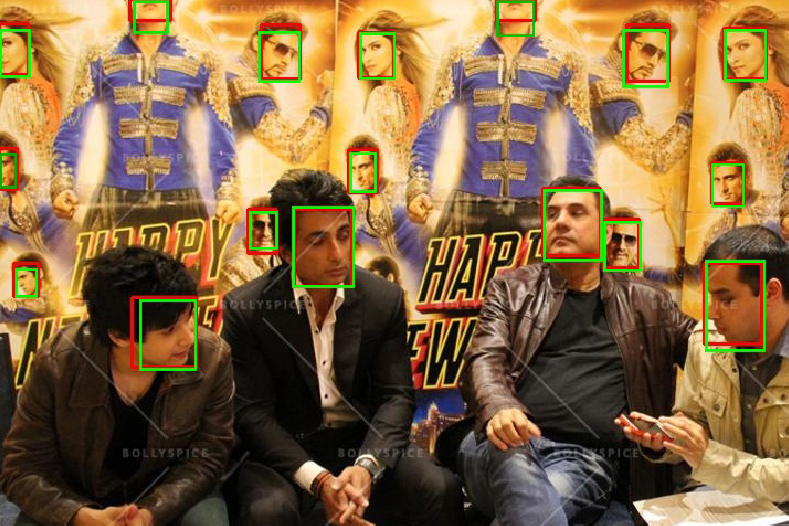}
		}				
	\end{minipage}
	\caption{Qualitative results on the WIDER FACE validation set. Red bounding boxes are the faces that annotated on the WIDER FACE validation dataset. Green bounding boxes represent the detection results. Best viewed in color. Please zoom in to see some small detections.}
	\label{fig:6}
\end{figure*}

\begin{figure*}
	\begin{minipage}{\linewidth}
		\centerline{
			\includegraphics[width=0.7in]{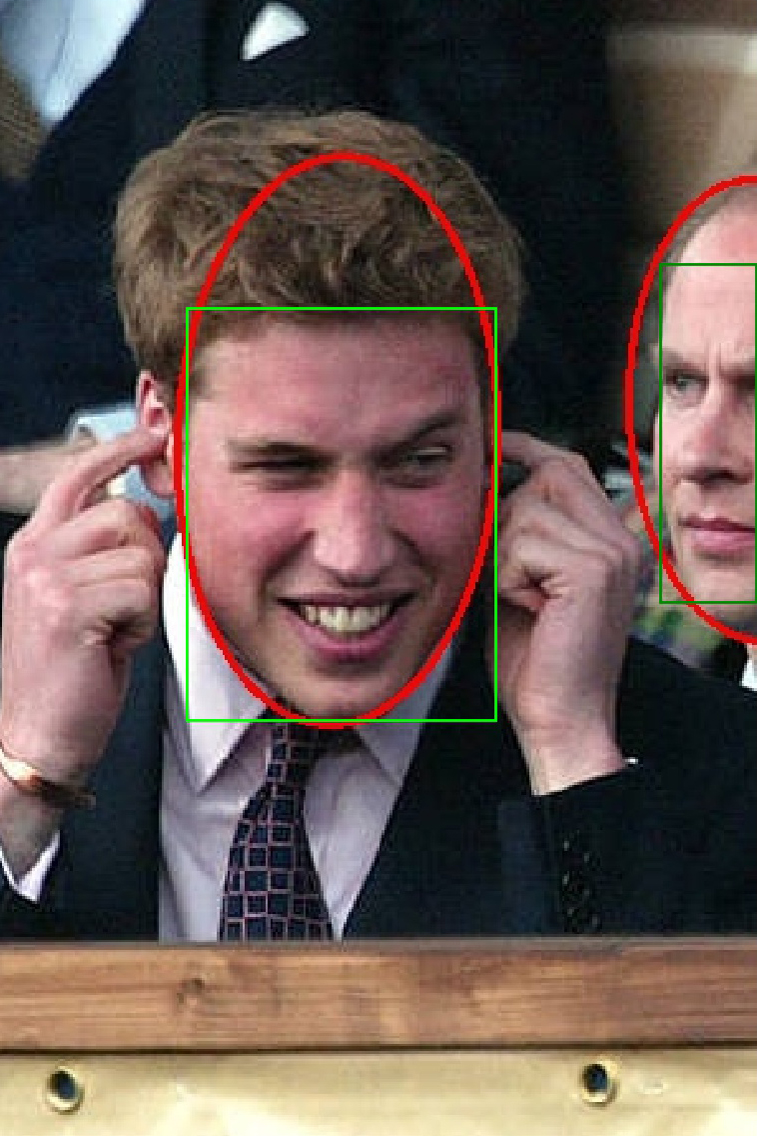}
			\includegraphics[width=0.7in]{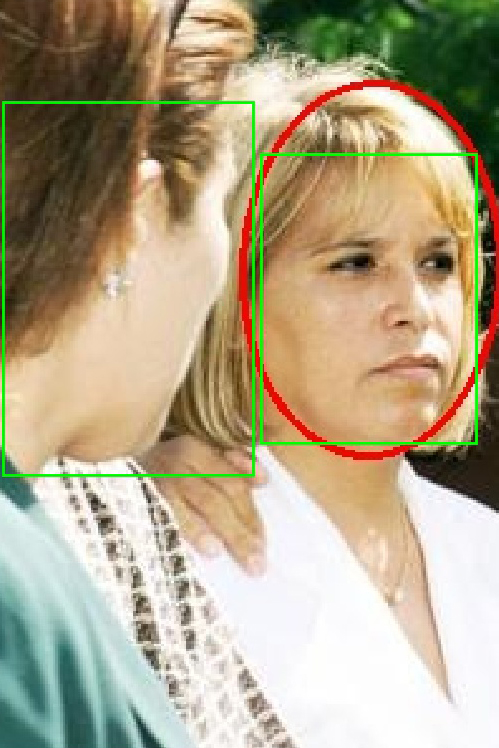}
			\includegraphics[width=0.7in]{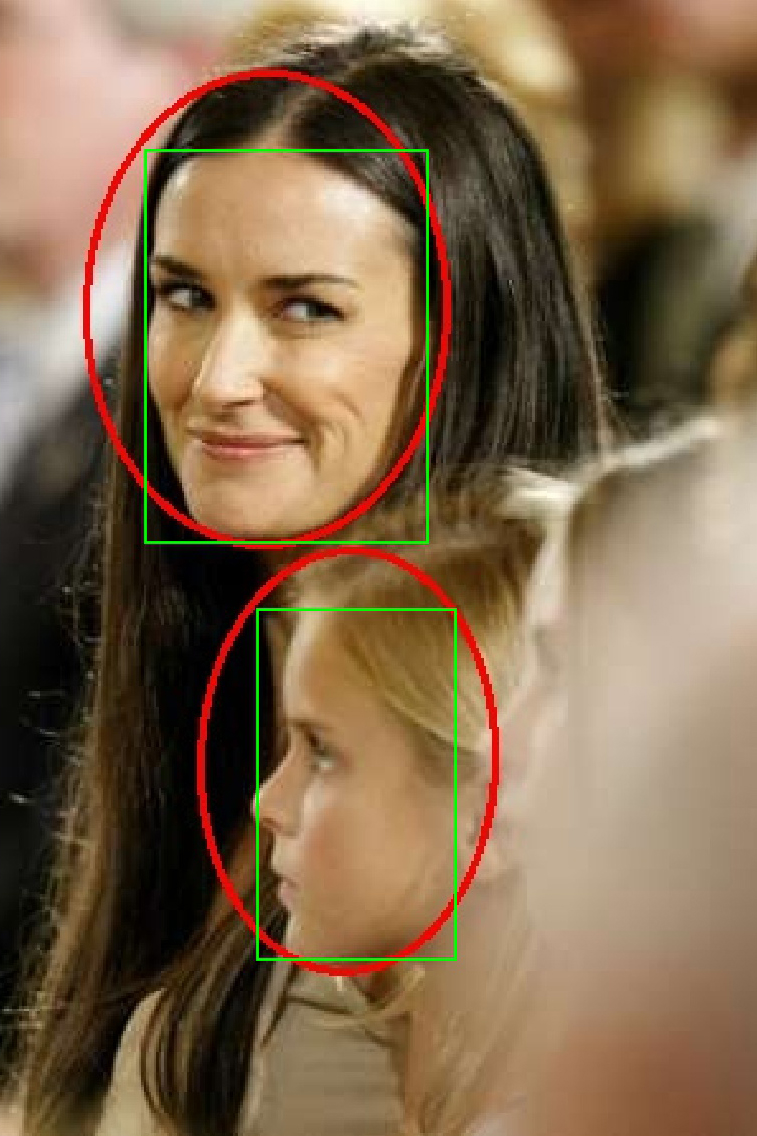}
			\includegraphics[width=0.7in]{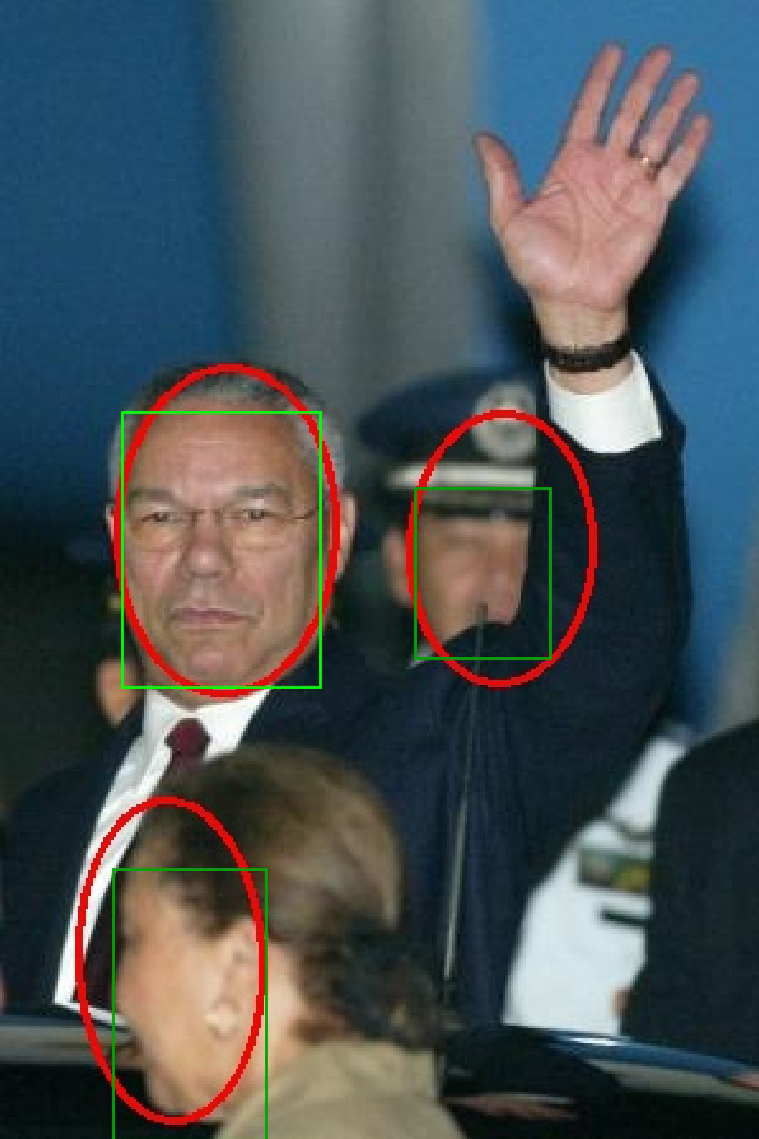}
			\includegraphics[width=0.7in]{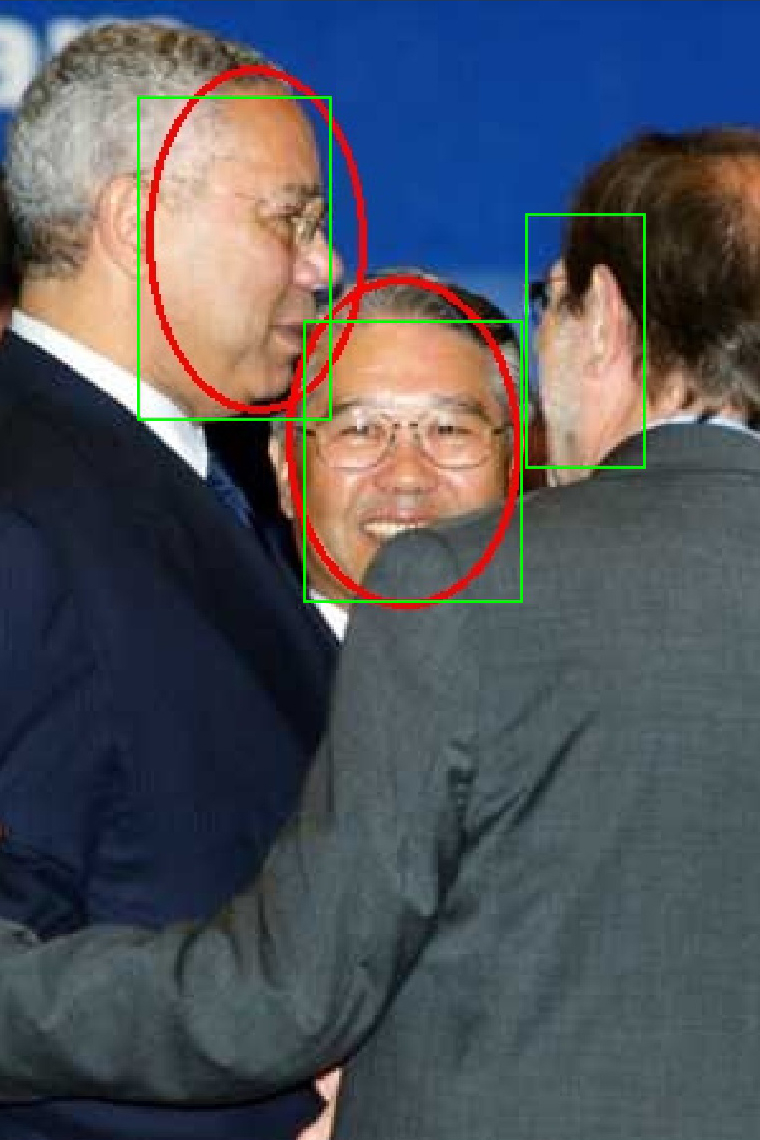}
			\includegraphics[width=0.7in]{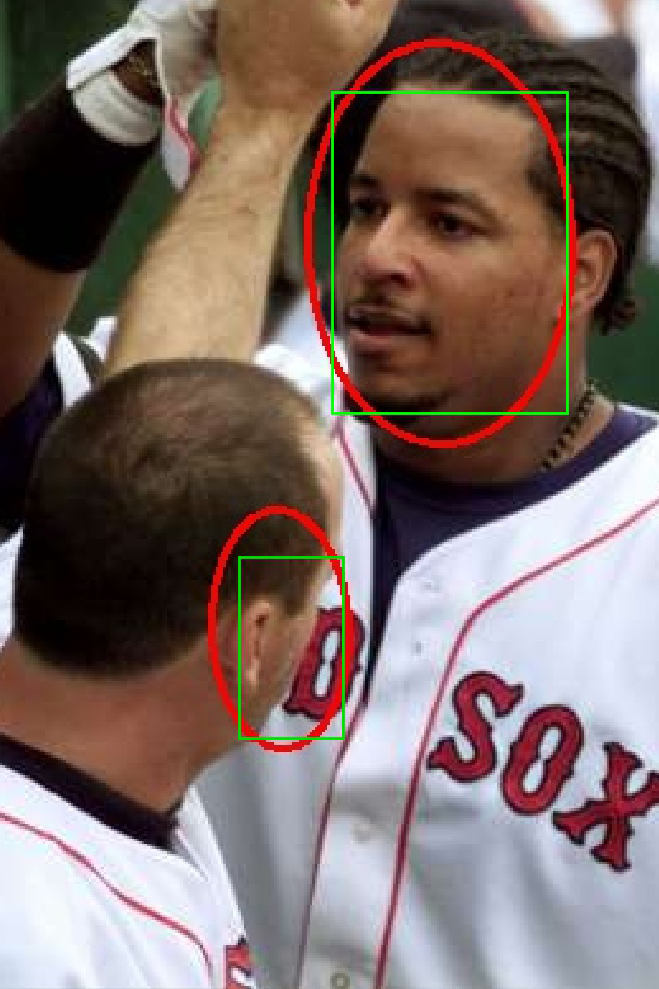}
		}
		\centerline{
			\includegraphics[width=0.7in]{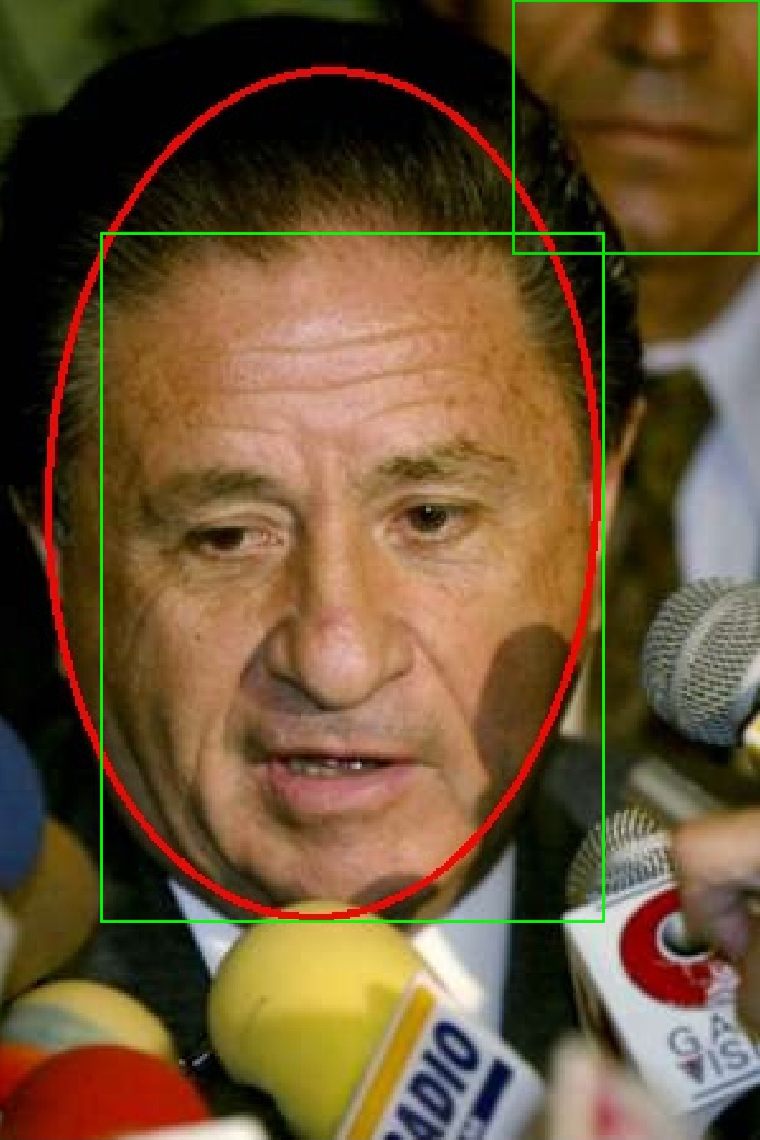}
			\includegraphics[width=0.7in]{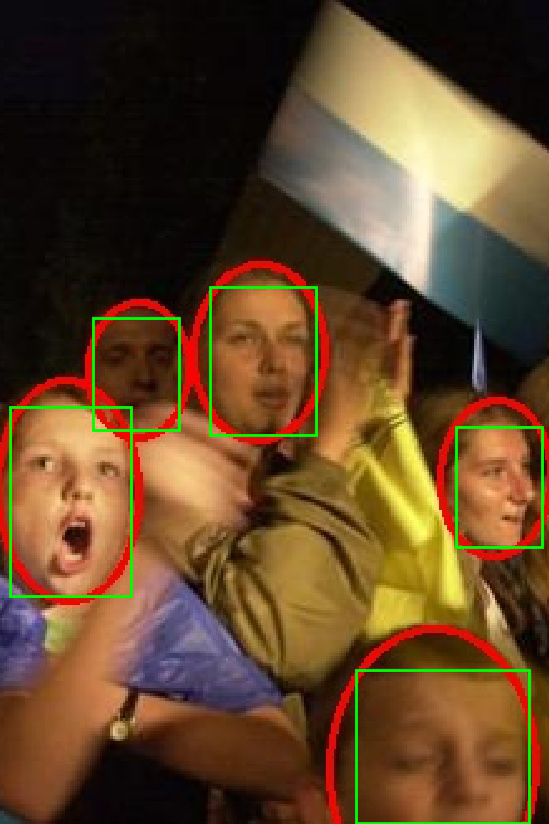}
			\includegraphics[width=0.7in]{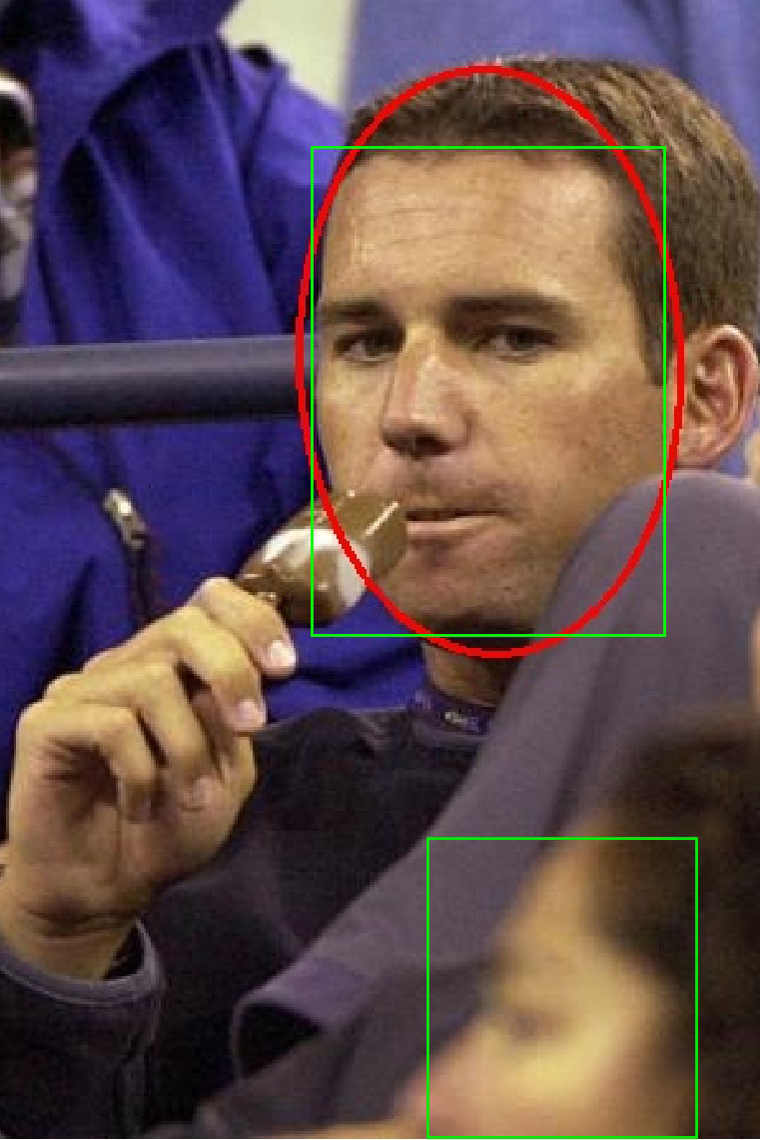}
			\includegraphics[width=0.7in]{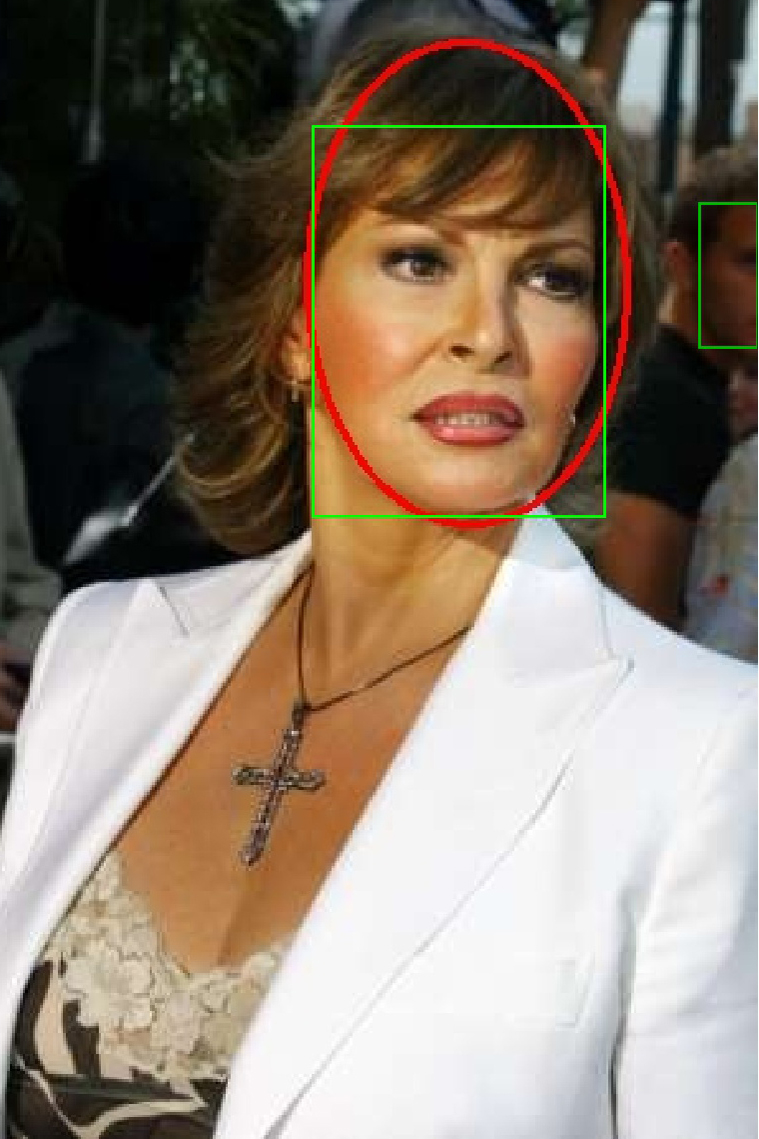}
			\includegraphics[width=0.7in]{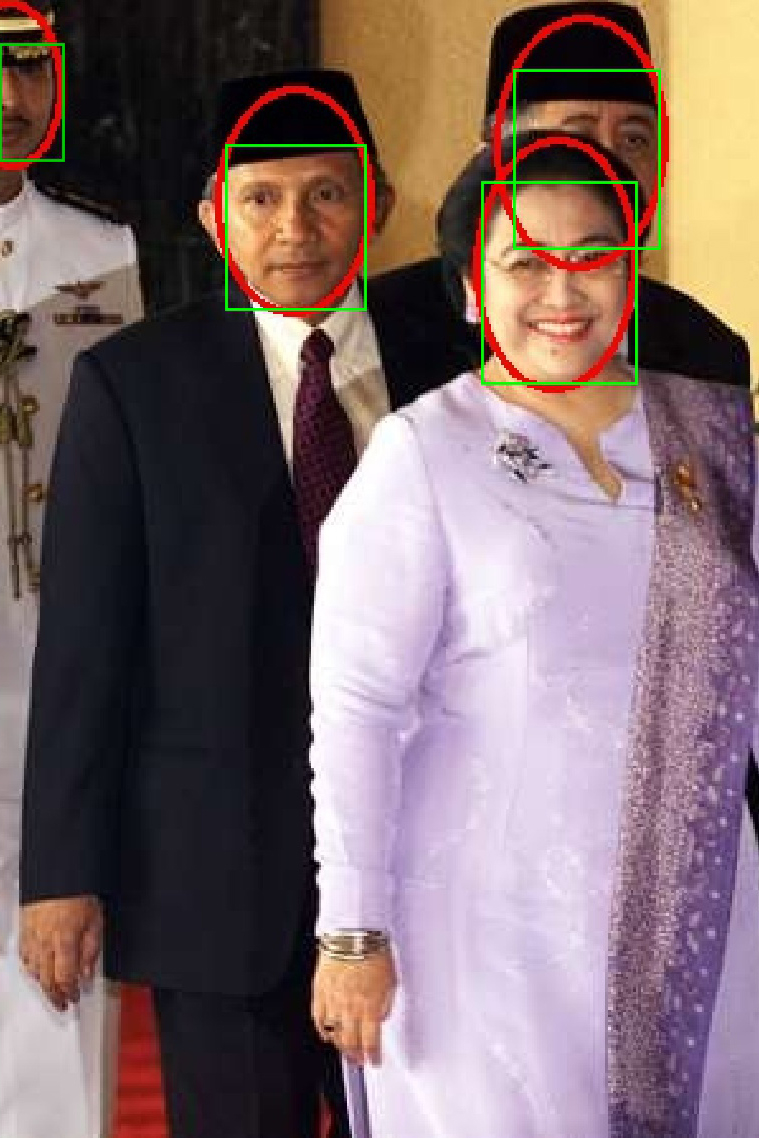}
			\includegraphics[width=0.7in]{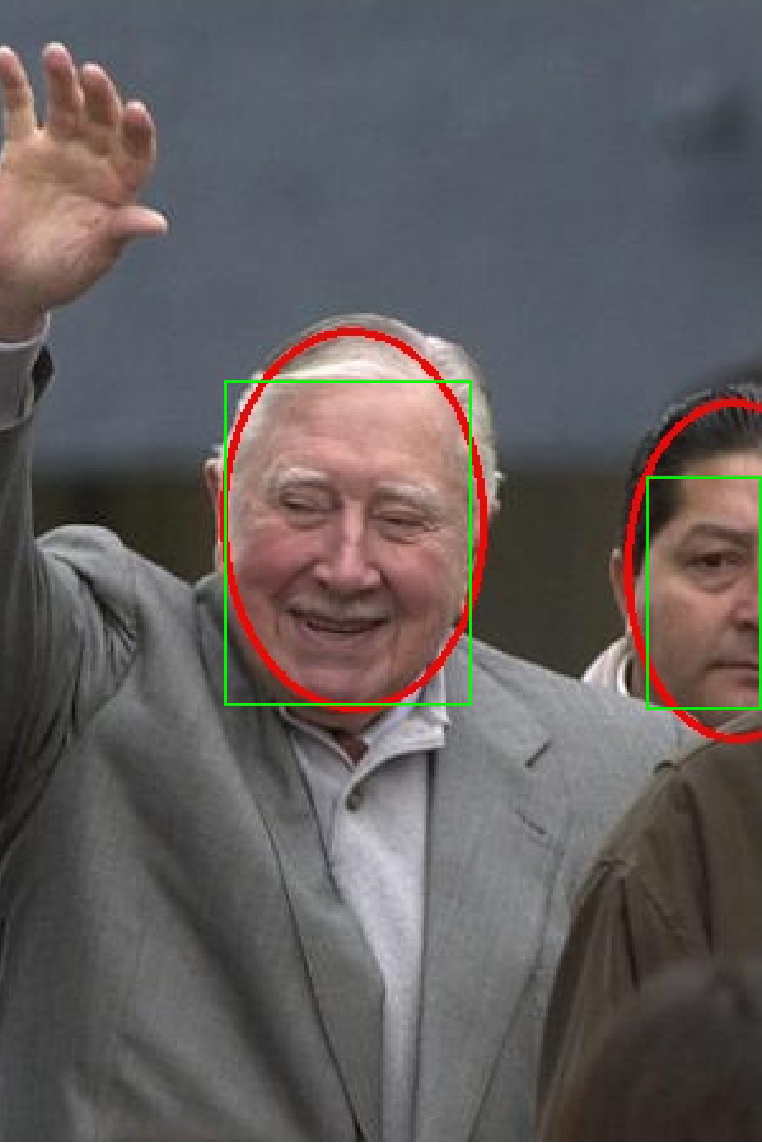}
		}				
	\end{minipage}
	\caption{Qualitative results on the FDDB dataset. Red bounding ellipses are the faces that FDDB labeled; Green bounding boxes are the detection results. Best viewed in color. Please zoom in to see some small detections.}
	\label{fig:7}
\end{figure*}

\begin{figure}
	\centering
	\includegraphics[width=4.5in,height=2in]{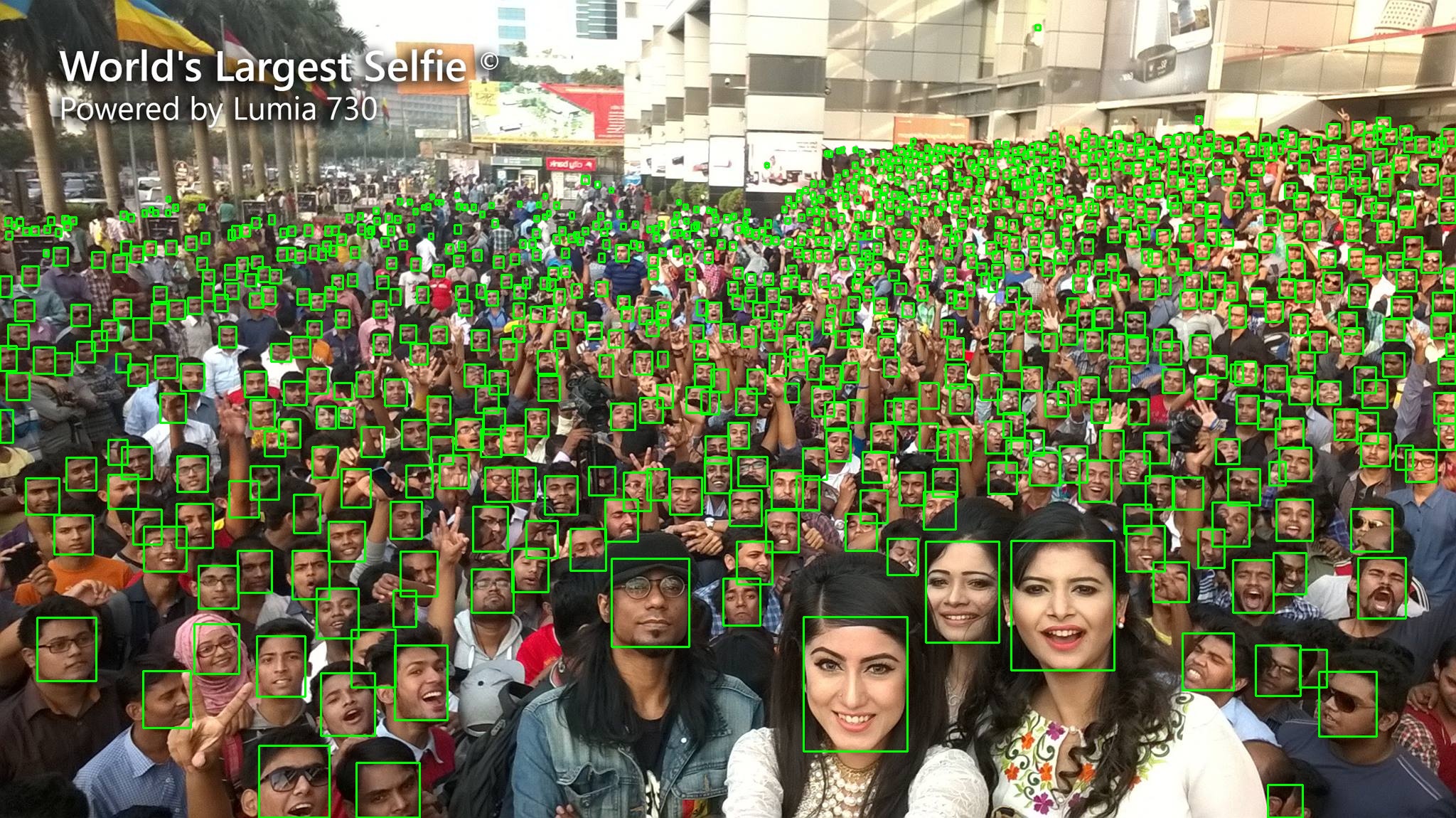}
	\caption{The qualitative result in world's largest selfie. Our method successfully find 911 faces out of the reported 1000 faces in the above image. Best viewed in color. Please zoom in to see some small detections.}
	\label{fig:2}
\end{figure}

Fig. \ref{fig:6} shows some detection results of our proposed method on the WIDER FACE validation dataset. Our method is able to detect faces with different AR, especially for extreme AR faces. The detection results for extreme pose faces are shown in the first row of Fig. \ref{fig:6}. Besides, our method can also detect partial faces caused by occlusion (see the last row of Fig.\ref{fig:6}). Surprisingly, our detection model can capture some extra extrmeme AR faces which are missing labels.\\

\indent{} Fig. \ref{fig:7} shows some detection results generated by our detection model on the FDDB dataset. Benefit from excellent performance of our method in detecting extreme AR faces, we can find some more faces from human perspective but lack of labels on the FDDB dataset. Similarly, the detection results, which contain atypical pose and heavy occlusion faces, are presented in the first and last row of Fig. \ref{fig:7} separately.\\

\indent{} Fig. \ref{fig:2} is a qualitative result in world's largest selfie. Our method successfully find 911 faces out of the reported 1000 faces in the above image.
\section{Conclusions}
\label{sec5}
In this paper, we examine the failure of sampling positive anchors from extreme AR faces and identify that the max IoUs of these faces are still lower than fixed sampling threshold in SAM strategy. Motivated by this observation, both Wide Aspect Ratio Matching strategies and Receptive Field Diversity module are deployed in our method for the sake of better detecting faces with different aspect ratios. These two strategies make our model effective and robust to detect faces with diversified AR in unconstrained settings, especially for extreme AR faces. Extensive experiments demonstrate that our method outperforms most of the recently published face detectors and achieves promising performance on challenging face detection benchmarks like WIDER FACE and FDDB datasets.
\section{Acknowledgements}
\label{sec6}
The work was supported in part by the National Natural Science Foundation of China under Grant 61801190, in part by the National Key Research and Development Project of China under Grant 2019YFC0409105, in part by the Nature Science Foundation of Jilin Province under Grant 20180101055JC, in part by the Industrial Technology Research and Development Funds of Jilin Province under Grant 2019C054-3, in part by the "Thirteenth Five-Year Pla" Scientific Research Planning Project of Education Department of Jilin Province (JKH20200678KJ,JJKH20200997KJ), and in part by the Fundamental Research Funds for the Central Universities, JLU.


%
%


%
%



\end{document}